\documentclass[twoside,11pt]{article}
\pdfoutput=1
\usepackage{jmlr2e}
\usepackage[utf8]{inputenc} 
\usepackage{booktabs}       
\usepackage{amsfonts,upgreek}       
\usepackage{nicefrac}       
\usepackage{microtype}      
\usepackage{color}
\usepackage{algorithm}
\usepackage{algpseudocode}
\usepackage{amssymb}

\usepackage{array,multirow}
\usepackage{subcaption}
\usepackage[export]{adjustbox}
\usepackage{mdframed,color}
\usepackage{float}

\usepackage{amsmath}
\usepackage{amssymb}
\usepackage{algorithm}
\usepackage{color,paralist}
\usepackage{comment}
\usepackage{float}
\usepackage{comment}
\usepackage{listings}
\usepackage{epsfig}

\usepackage{amsfonts}       
\usepackage{nicefrac}       
\usepackage{microtype}      
\usepackage{xcolor}
\usepackage{amsmath,amssymb}
\usepackage{algpseudocode}

\usepackage{tikz}
\usetikzlibrary{shapes,snakes,calc,fit,shadows}

\newcommand{\todo}[1]{}
\renewcommand{\todo}[1]{{\color{red} TODO: {#1}}}

\newcommand{\R}{\mathbb{R}}

\newcommand{\reg}{\mathcal{R}}
\newcommand{\dist}{\mathcal{D}}
\newcommand{\loss}{\text{loss}}
\newcommand{\score}{\mathcal{S}}
\newcommand{\dual}{\mathsf{d}}

\newcounter{obs}
\newtheorem{obsn}[obs]{Observation}

\definecolor{abhay1}{RGB}{233,243,245}
\definecolor{abhay2}{RGB}{63,127,147}
\tikzstyle{mybox} = [draw=red, fill=abhay1, thick,
    rectangle, inner sep=0.25pt, outer sep=0.25pt, inner ysep=0.75pt, inner xsep=1pt]
\tikzstyle{fancytitle} =[fill=abhay2, text=white]


\newcommand{\cfbox}[1]{%
	\adjustbox{cfbox=#1}%
}

\definecolor{light-gray}{gray}{0.95}
\newcommand{\code}[1]{\colorbox{light-gray}{\texttt{#1}}}

\tikzset{note/.style={rectangle callout, rounded corners,fill=gray!20,drop shadow,font=\footnotesize}}    

\newcommand{\tikzmark}[1]{\tikz[overlay,remember picture] \node (#1) {};}    

\newcounter{image}
\makeatletter
\newenvironment{btHighlight}[1][]
{\begingroup\tikzset{bt@Highlight@par/.style={#1}}\begin{lrbox}{\@tempboxa}}
{\end{lrbox}\bt@HL@box[bt@Highlight@par]{\@tempboxa}\endgroup}

\newcommand\btHL[1][]{%
  \begin{btHighlight}[#1]\bgroup\aftergroup\bt@HL@endenv%
}
\def\bt@HL@endenv{%
  \end{btHighlight}%
  \egroup
}
\newcommand{\bt@HL@box}[2][]{%
  \tikz[#1]{%
    \pgfpathrectangle{\pgfpoint{0pt}{0pt}}{\pgfpoint{\wd #2}{\ht #2}}%
    \pgfusepath{use as bounding box}%
    \node[anchor=base west,rounded corners, fill=green!30,outer sep=0pt,inner xsep=0.2em, inner ysep=0.1em,  #1](a\theimage){\usebox{#2}};
  }%
  \tikzmark{a\theimage}
 \stepcounter{image}
}
\makeatother

\usepackage{enumerate}
\usepackage{times}  
\usepackage{helvet} 
\usepackage{courier}  
\usepackage[hyphens]{url}  
\usepackage{graphicx} 
\usepackage{graphicx}  
\title{Optimizing Nondecomposable Data Dependent Regularizers via Lagrangian Reparameterization offers Significant Performance and Efficiency Gains}
\author{\centering Sathya N. Ravi \quad  Abhay Venkatesh \quad Glenn Moo Fung \quad Vikas Singh\\
	{\tt\footnotesize \centering  sathya@uic.edu, abhay.venkatesh@gmail.com, gfung@amfam.com, vsingh@cs.wisc.edu} }
\begin{document}
	
	\maketitle
	
	{\color{black}\begin{abstract}
  Data dependent regularization is known to benefit a wide variety of problems in machine learning. Often,
  these regularizers cannot be easily decomposed into a sum over a finite
  number of terms, e.g., a sum over individual example-wise terms. The $F_\beta$ measure, Area under the ROC curve
  (AUCROC) and Precision at a fixed recall (P@R)
  are some prominent examples that are used in many applications. We find that for most medium to large sized datasets, scalability
  issues severely limit our ability
  in leveraging the benefits of such regularizers. Importantly, the key technical impediment
	despite some recent progress   is that, such objectives remain difficult to optimize
  via backpropapagation procedures.
  While an efficient 
  general-purpose strategy for this problem still remains elusive, in
  this paper, we show that for many data-dependent nondecomposable regularizers that are relevant in applications,
  sizable gains in efficiency are possible with minimal code-level changes; in other words,
  no specialized tools or numerical schemes are needed.
  Our procedure involves a reparameterization
  followed by a partial dualization -- this leads to a formulation that has provably cheap
  projection operators. We present a detailed analysis of runtime and convergence properties of our algorithm.
  On the experimental side, we show that a direct use of our scheme significantly improves 
  the state of the art IOU measures reported for MSCOCO Stuff segmentation dataset. 
\end{abstract}
}
	\section{Introduction}
\label{sec:intro}
\begin{figure*}[t]
	\centering
	\includegraphics[width=\linewidth]{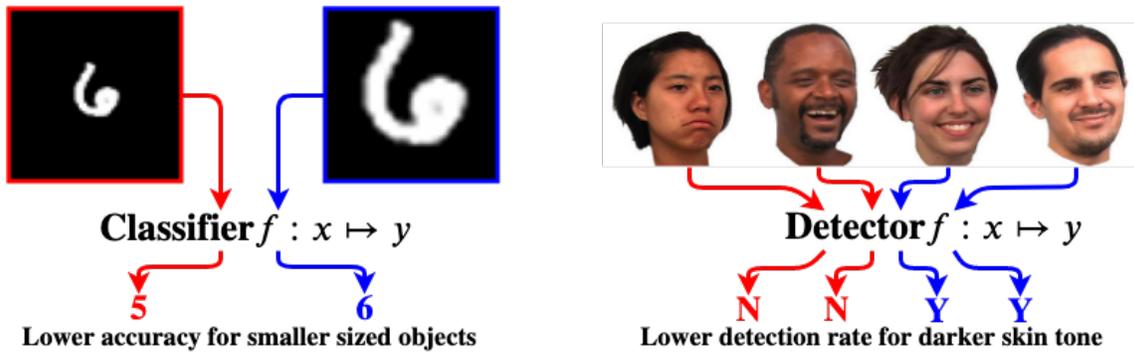}
	\caption{We illustrate the need for data dependent regularization. The performance of an image classifier (left) and face detector (right) can vary significantly depending on the size (small objects get misclassified) and skin tone  (weak or no detection power for darker skin) of the {\em data}.  A data  dependent term can explicitly make the classifier/detector to behave similarly for various subgroups in the population.}
	\label{fig:mnist_ex}
\end{figure*}
Data dependent regularization is a mature and effective strategy for many problems
in machine learning. 
In semi-supervised learning \cite{corduneanu2006data},
the marginal distribution of the examples may serve to influence 
the estimation of the conditional distribution and in SVMs, one could
optimize the maximum {\em relative} margin based on the data distribution, 
rather than the {\em absolute} margin. 
In modern deep learning, data-dependent regularization
is routinely used in both explicit and implict ways. 
A regularizer can take the form of certain
geometric or clustering-type constraints \cite{lezama2018ole,Zhu_2018_CVPR}
on the representations that are learned by the network -- 
using the distribution overlap between different classes \cite{rippel2015metric} or 
seeking decorrelated autoencoder latent codes \cite{cheung2014discovering}.
On the other hand, artificial corruption of the data 
is also a form of regularization --  dropout
method induces a regularization  on 
the Fisher information matrix of the samples 
\cite{wager2013dropout}.
More recently, the results in \cite{mou2018dropout} offer a nice treatment of the form of data-dependent
regularization that emerges from popular methods such as batch normalization
and AdaGrad. 

{\bf From Decomposable to Nondecomposable data-dependent regularizers.} A number of data-dependent regularizers described above
can be written as a sum of 
individual example-wise estimates of the regularizer. This setting is desirable because in order to run a SGD type algorithm, we simply pick a random sample to get an unbiased estimate of the gradient.
But a number of application domains often necessitate a regularization criteria that may not 
decompose in this manner. In such settings, a random sample (minibatch) of the dataset  does not provide us an unbiased gradient -- biased gradients are known to adversely impact both the qualitative and quantitative performance of the training procedure, see \cite{chen2018stochastic}.

{\bf Why are Nondecomposable regularizers relevant?} Consider the situation where we would like to ensure that the performance of a statistical model is uniformly good over groups induced via 
certain protected attributes (such as race or gender), see Figure \ref{fig:mnist_ex}.
Or alternatively, we want that
when updating an algorithm in a manufacturing process, the new system's behavior should mostly remain
similar with respect to some global measures such as makespan \cite{limentani2005beyond}. \cite{pmlr-v97-cotter19a} shows applications of various shape constraints that are associated with set functions.
And finally, when pooling datasets from multiple sites, global characteristics of Precision-Recall should be (approximately) preserved across sites \cite{zhou2017can}. Essentially, these applications suggest that the performane of a model in expectation (on the entire population), does not automatically guarantee that the model will perform well on specific subgroups. Motivated by these issues encountered in various
real world problems, recently \cite{cotter2018optimization} presents a comprehensive study of the computational aspects of  learning problems with {\em rate constraints} -- there,
the constrained setting is preferred (over penalized setting) due to several reasons including
its ease of use for a practitioner. The authors
show that for a general class of constraints, a proxy-Lagrangian based method must
be used because the Lagrangian is not optimal.
This raises the question whether there exist a broad class of data-dependent {\em nondecomposable}
functions for which the regularized/penalized formulation based on {\em standard} Lagrangian schemes may, in fact,
be effective and sufficient. In this paper, we will address this question with simple examples shortly.

{\bf Examples in statistics.} Nondecomposable regularization, in the most general sense, has also
been studied from the statistical perspective, and is
often referred to as \emph{shape} constraints \cite{groeneboom_jongbloed_2014}. The need for
shape constraints arise in clinical trials and
cohort studies of various forms in the competing risk model,
formalized using local smooth function theory, see Chapter 5 in \cite{daimon2018introduction} and \cite{chenxi2012smooth}.
While several authors have studied specific forms of this problem, 
the literature addressing the {\em computational}
aspects of estimation schemes involving such regularizers is sparse, and even for 
simple objectives such as a sparse additive model, we find that results have appeared
only recently \cite{yin18convex}.
Leveraging these ideas to train richer models of the forms
that we often use in modern machine learning, establishing their convergence properties, and
demonstrating their utility in real world applications is still an open problem.
 
\begin{figure*}[!t]
	\centering
	\includegraphics[width=\linewidth]{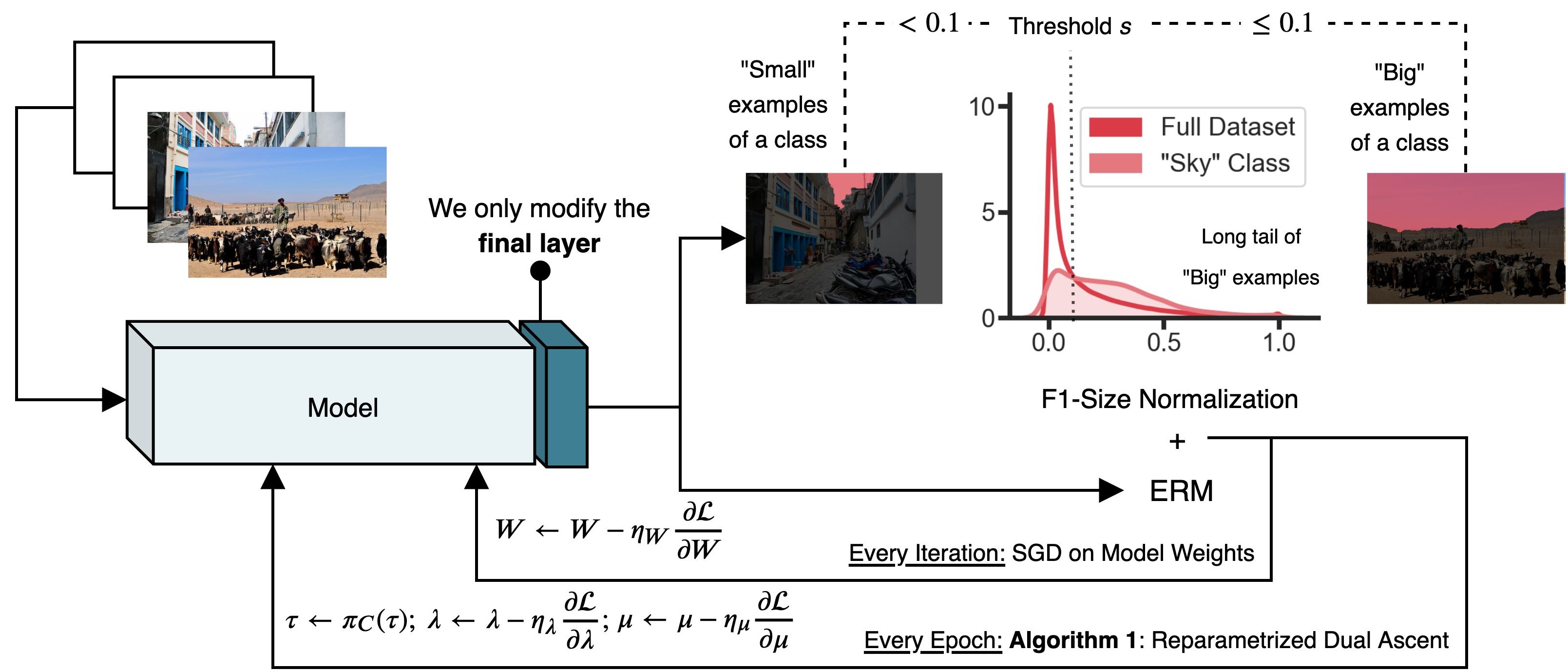}
	\caption{Our proposed training pipeline for optimizing nondecomposable $\score-$Measures.}
	\label{fig:our_model}
\end{figure*}

{\bf Our Contributions.}  We first {\em reparameterize} a  broad class of  nondecomposable data-dependent regularizers into a form that can be efficiently optimized using first order methods. Interestingly, this reparameterization
naturally leads to a Lagrangian based procedure where existing SGD based methods can be employed with little to no change, see Figure \ref{fig:our_model}.
While recent results suggest that optimizing nondecomposable data-dependent regularizers may be 
challenging \cite{cotter2018optimization}, our development shows that a sizable subclass of such regularizers indeed admit simple solution schemes.
Our overall procedure comes with convergence rate guarantees and optimal per-iteration complexity.
On the MSCOCO stuff segmentation dataset, we show that a direct use of this technique yields significant improvements
to the state of the art, yielding a mean IoU of 0.32. These results have been submitted to the leaderboard. 

	\section{Preliminaries}
{\bf Basic notations.} We assume that the training data is given as pairs of $(x,y)\sim \dist$ where the joint distribution $\dist$ is unknown.
Here, $x,y$ are random variables that represent examples (e.g., images) and labels respectively. We make
no assumptions on $x$ and $y$ in that the marginal distributions of $x$ and $y$ can be discrete or continuous. Our goal is to
estimate a function $f:x\mapsto y$ that achieves the minimum error measured using a specified loss function on the 
empirical/observed samples. We will  use $W=\{w_1,w_2,\dots,w_{l+1}\}$ to represent the trainable parameters of a  feedforward neural network with $l+1$ layers,
with nonlinear activation functions. The output $W(x)$ of the function computed by the neural network $W$ may be used
to construct the classification or regression function $f$.

{\bf Nondecomposability.} Now, suppose there exists a function $\varphi$ such that $\varphi\circ \dist=:s$ is a  random variable called ``{\em shape}'' in
a similar sense as described above \cite{groeneboom_jongbloed_2014}.
Our notation is suggestive of the fact  that $s$ is {\bf nondecomposable}, i.e., the value $s_i$ for individual datapoint/example $i$ may
depend on the {\em entire} (empirical) distribution of $\dist$.  Moreover, this  implies that without any further assumptions,
{\em any}  loss function $\reg(\cdot) $ on $W(x)$ used to learn a model $g$ to predict $s$ cannot be represented as a decomposable function with respect
to the training data $x,y$, see \cite{narasimhan2014statistical,sanyal2018optimizing}.

{\bf } As briefly noted in Section \ref{sec:intro}, nondecomposability poses unique
challenges in designing algorithms for
learning models with a large number of parameters. In most cases,  existing backpropagation based techniques may {\em not}
be directly applicable and require careful modifications \cite{zhang2018faster}.
In other cases,  they may require significant computational overhead due to expensive projection operations \cite{kar2014online}.
Before we present our model, we now briefly describe two fundamental tasks that are, by and large, the most frequently used modules in vision based pipelines
to see the benefits of nondecomposability. 
{\color{black}
  \paragraph{Example 1.} Object Detection seeks to simultaneously localize and classify objects
  in an image using a bounding box for each object \cite{zhao2019object}. 
Naturally, since the bounding boxes are rectangular, the {\em data-dependent}
regularizer $s$ is fully determined by the size of the objects. In other words, $s$ here becomes the area of the bounding box of the object.
Here, we may use the knowledge of $s$ to learn $f$ that can perform equally well over
{\em all} sizes present in the training data.
But even though we may simply use class activation maps to compute
the empirical distribution of $s$ \cite{zhou2016learning},
the loss functions $\reg(s)$ that are commonly used are {\bf nondecomposable}
as we will see shortly. 

\paragraph{Example 2.} Semantic Segmentation seeks to assign
{\em each} pixel of an image to a class \cite{chen2018encoder}.
Recent works suggest that in order to train a model $f$,
we may choose a model whose complexity strongly
depends on the number of pixels in the images, see \cite{liu2018recent} for a recent survey.
Unfortunately, these methods use up/down-sampling and so the learned representations
do not directly offer immunity to variations in how much of the image is occupied
by each class (i.e., size). 

\begin{remark}
  {\color{black}  We use these two examples
    to illustrate the applicability of our developments -- since the
    use case corresponds to size, 
    we may assume that the ``shape'' random variable is {\em discrete}.
    To that end, we will use   $\score$ to denote the {\em countable scoring set}
    or the support of $s$ with $|\score|<\infty$.  }
\end{remark}
}

	\subsection{Incorporating $\score-$measures for Optimization}\label{sec:model} 

Let us write down a formulation which incorporates our shape regularizer. The
objective function is a sum of two terms: \begin {inparaenum} [\bfseries (1)] 
\item a {\em decomposable} Empirical Risk Minimization (ERM) term to learn the optimal function which maps $x$ to $y$ and
\item a {\em nondecomposable, data dependent} $\score-$measure regularizer term for $s$ from $x$. 
\end {inparaenum} In particular, for a fixed $\alpha>0$, we seek to solve the following optimization problem,\begin{align}
\min_{W} \quad \overbrace{\frac{1}{N}\sum_{i=1}^{N} \loss\left(W;x_i,y_i\right)}^{\text{ERM}} ~+~ \alpha \overbrace{ \reg\left(W;\varphi\circ \hat{D}\right)}^{\score-\text{Measure}},\label{eq:obj_gen}
\end{align}
where  $\hat{D}$ represents the empirical distribution, $x_i\in\R^d$ and $y_i\in\R^k$ with $i=1,\dots,N$ denoting training data examples.
We let $\loss(\cdot)$ to be any standard loss function such as a cross entropy loss, hinge loss and so on.
To keep the presentation and notations simple but nonetheless illustrate our main
algorithmic developments, {\bf we will assume that the $\score-$Measure is given by $F_{\beta}$ metric} \cite{lipton2014optimal} while noting that 
the results apply directly to other measures such as R@P, P@R and can
be easily generalized to other nondecomposable metrics {\color{black}such as AUCROC, AUCPR}, using the Nystr\"om method with no additional computational overhead \cite{eban2016scalable}.
\begin{remark}
  For simplicity, we will suppress the dependence of $f$ computed by the parameters $W$ in our objective function \eqref{eq:obj_gen}.
  For example, if $f$ is represented using a deep neural network with weights $W$, then both terms in \eqref{eq:obj_gen} may be nonconvex. 
\end{remark}

	\section{Reparameterizing $\score-$Measures}
Since any backpropagation based procedure can easily handle decomposable terms,
let us ignore the ERM term in our objective function \eqref{eq:obj_gen} for the moment, and focus on the second term. 
Specifically, in this section, we will focus on the $F_{\beta}$ metric as a placeholder for $\score-$measures. 
and show that there is a reformulation which will enable a backpropagation based procedure to solve our problem \eqref{eq:obj_gen}.
For simplicity, let us assume that
$s\in\{+1,-1\} $ and $\beta=1$. {\color{black}The
appendix shows how to extend our approach to any finite $|\score|$ and $\beta\in (0,1]$} noting that 
in various applications in Section \ref{sec:exps}, $ | \score| =2$ suffices and is already quite effective (e.g., the learned model works whether $s_i$ is $+1$ or $-1$). 

The starting point of our technical development is the following result,

\begin{obsn}[Restated from \cite{eban2016scalable}]
  \label{obs:eban}
  If $f$ is linear, i.e., $y=W\cdot x$, we can represent the $\score-$measure term as a Linear Program (LP) as shown in equation (13) in \cite{eban2016scalable}.
\end{obsn}

{\bf Moving forward from Observation \ref{obs:eban}}. In principle, one could use standard LP solvers to solve the LP in \cite{eban2016scalable}. But this requires
instantiating the constraint matrix (linear in the number of samples). This is impractical -- the alternative 
in \cite{eban2016scalable} is to use an iterative ascent method. In general this will also require a
projection scheme, e.g., solving a QP. While more attractive than using an off-the-shelf method,
both options may need cubic time.
  \begin{center}
    \begin{tikzpicture}
	\node [mybox] (box){%
		\begin{minipage}{\linewidth}
		  {\footnotesize
                    \begin{subequations}
                      \label{eq:prob1}
		    \begin{align}
		      \underset{\phi,\tau,w,\epsilon}\min & \quad n\epsilon + \sum_{i\in S^+} \tau_i +\sum_{i\in S^-}\phi_i\label{eq:obj}\\
		      \text{s.t.} & \quad\tau_i\leq \epsilon, \tau_i\leq w_{l+1}\cdot a_l(x_i); \,\forall~i\in  S^+ \label{eq:cons_positive}\\
		      \ & \quad\phi_i\geq 0\,, \phi_i\geq \epsilon+ w_{l+1}\cdot a_l(x_i)\label{eq:cons_negative}; \,\forall i \in S^- \\
		      \ & \quad\sum_{i\in S^+} \tau_i =1\, \quad\epsilon\geq 0\,.\label{eq:cons_hyperplane}
		    \end{align}
                    \end{subequations}
		}
		\end{minipage}
	};
	\node[fancytitle, right=10pt] at (box.north west) {\footnotesize Problem \eqref{eq:prob1}: Slightly adapted form of LP in \cite{eban2016scalable}};
	\end{tikzpicture}%
  \end{center}

Further, this discussion of the computational issues only pertains to a {\em linear} objective -- the setting
expectedly becomes much more challenging for non-convex objectives common in deep learning architectures.

\subsection{Simplify and Reparameterize}
We will first generalize the construction in \cite{eban2016scalable} to be more amenable to non-linear functions that one often
estimates with DNN architectures.
Note that since our data-dependent $\score-$measures are specified on the input distribution of examples or representations derived
from the transformations induced by the network $W$, we may denote representations in general as $a_l(x_i)$: in
other words, the $l-$th layer provides us a {\em representation} of example $i$.
We define $S^+:=\{i:s_i=+1\} $ (and similarly $S^-$) with $|S^+|=n$, and $n+|S^-|=N$,
calculated using only the training data samples $X:=\{x_i\},Y:=\{y_i\}, i=1,...,N$.
{We will still focus on the nondecomposable term but
bring in the DNN loss when we describe the full algorithm. }

  \begin{center}
    \begin{tikzpicture}
	\node [mybox] (box){%
		\begin{minipage}{\linewidth}
		  {\footnotesize
                    \begin{subequations}
                      \label{eq:prob2}
			\begin{align} 
			\min_{\tau,w,\epsilon} & \quad n\epsilon + \sum_{i\in S^-}\max(0,\epsilon+w_{l+1}\cdot a_l(x_i)) \label{eq:obj_1}\\
			\text{s.t.} & \quad\tau_i\leq \epsilon, \tau_i\leq w_{l+1}\cdot a_l(x_i)\label{eq:cons_positive_1}; \forall~i\in S^+\\
			\ & \quad\sum_{i\in S^+} \tau_i =1\, \quad\epsilon\geq 0\,.\label{eq:cons_hyperplane_1}
			\end{align}
                    \end{subequations}  
		}
		\end{minipage}
	};
	\node[fancytitle, right=10pt] at (box.north west) {\footnotesize Problem \ref{eq:prob2}: a simplified LP};
	\end{tikzpicture}%
  \end{center}
\paragraph{Eliminating redundancy in Problem \eqref{eq:prob1}.} First, notice that we can eliminate the $\sum_{i\in S^+}\tau_i$ term from the objective function in  \eqref{eq:obj} since any feasible solution has to satisfy the first constraint in \eqref{eq:cons_hyperplane}.
Second, using the definition of hinge loss, we can eliminate the $\phi_i$ terms in the objective
by replacing $\phi_i$ with $\max(0,\epsilon+w_{l+1}\cdot a_l(x_i))$. This rearrangement leads
to a LP in Problem \eqref{eq:prob2}. \begin{remark}The choice of $\phi$ to be the hinge loss is natural in the formulation of $F_{\beta}$ metric. Our Algorithm \ref{algorithm1} can easily handle other commonly used loss  functions such as Square, Cross Entropy, and Margin. 
\end{remark}

\paragraph{Reparameterization Via Partial Dualization.}
Problem \eqref{eq:prob2} has a objective which is expressed as a sum over finite number of terms.
But overall, the problem remains difficult because the 
constraints \eqref{eq:cons_positive_1} are nonlinear and problematic. 
It turns out that a {\em partial dualization} -- for only the problematic constraints --
leads to a model with desirable properties. 
Let us dualize {\em only} the $\tau_i\leq w_{l+1}\cdot a_l(x_i)$ constraints in \eqref{eq:cons_positive_1} using $\lambda_i$,
and the equality constraint in \eqref{eq:cons_hyperplane_1} using $\mu$, where $\lambda_i$ and $\mu$ are the dual variables.
This yields
the Lagrangian,\begin{align}
&L
:=n\epsilon +\sum_{i\in S^-}\max(0,\epsilon+w_{l+1}\cdot a_l(x_i))\notag\\&+ \mu\left(\sum_{i\in S^+}\tau_i-1\right)+\sum_{i\in S^+} \lambda_i\left(\tau_i-w_{l+1}\cdot a_l(x_i)\right).\label{eq:lagrangian}
\end{align}
We can denote the set $C:=\{(\tau_1,\tau_2,...,\tau_{n},\epsilon):\tau_i\leq \epsilon,\epsilon\geq0,i=1,...,n\}$.
With this notation,
{our final optimization problem for the $\score-$measure can be} equivalently written as,\begin{align}
\max_{\lambda\geq 0,\mu}\min_{(\tau,\epsilon)\in C,W} L(\tau,w,\epsilon,\lambda,\mu).\label{eq:fin_lag}
\end{align}
We will shortly describe some nice properties of this model. 

{\bf Relation to ``Disintegration''.}
While procedurally the derivation above is based on numerical optimization, the form in \eqref{eq:fin_lag} is
related to {\em disintegration}. Disintegration  
is a measure theoretic tool that allows us to rigorously define conditional probabilities \cite{faden1985existence}.
As an application, using this technique, we can
represent any (also, nondecomposable) measure defined over the graph of a convex function
using (conditional) measures over its {\em faces} \cite{caravenna2010disintegration}.
Intuitively, at the optimal solution of \eqref{eq:fin_lag}, the dual multipliers $\lambda_i$
can be thought of a representation of the ``disintegrated'' $\score-$measure,
as seen in the last term in \eqref{eq:lagrangian}. In words, if we can successfully disintegrate the $\score-$measure, then there exists a loss function that can be written in the weighted ERM form such that the optimal solutions of the resulting ERM and that of the nondecomposable loss coincide. 
But in general, Lagrangians are difficult to use for disintegration because
\begin{inparaenum}[\bfseries]
\item $\lambda_i'$s need not sum to one; and
\item more importantly, when the optimal solution is in the {\em interior} of the feasible set in
  Prob. \eqref{eq:prob2}, then $\lambda_i= 0 \, \forall i$ by complementarity slackness.
\end{inparaenum}
This means that
the decomposition provided by the $\lambda_i'$s need {\em not} be a nontrivial (probability) measure
in a rigorous sense.
	
	%
	%
	

\subsection{Core benefit of Reparameterization: Efficient Projections will become possible}
The previous section omitted description of any real benefits of our reparameterization. 
We now describe the key computational benefits. 
\paragraph{Can we perform fast Projections? Yes, via Adaptive Thresholding.}
Observe that for the formulation in \eqref{eq:fin_lag} to offer any meaningful advantage,
we should be able to show that the Lagrangian can indeed
be efficiently minimized with respect to $\tau,\epsilon,W$.
This will crucially depend on whether we can project on to the set $C$ defined above. We now
discuss that this holds for different batch sizes. 

\begin{figure}[!t]
	\begin{minipage}{\textwidth}
		\begin{algorithm}[H]
			\centering
			\caption{Reparameterized Dual Ascent for solving  \eqref{eq:fin_lag}}\label{algorithm1}
			\begin{algorithmic}[1]
				\State Input: $X,Y,S,c\in(0,1)$ trainable parameters: $W,\uptau$                   
				\State Initialize variables $W,\uptau,T,\text{Epochs},\eta_d=c/e$
				\For{$e= 1,\dots, \text{Epochs}$}
				\For{$t= 1, \dots, T$}
				\State SGD on ERM  in \eqref{eq:obj_gen} + Lagrangian \eqref{eq:lagrangian} 
				\EndFor
				\State $\uptau \gets \Pi_{C}(\uptau)$ using Algorithm \ref{algorithm2}
				\State $\lambda\gets \lambda + \eta_d(\tau - w_{l+1}\cdot a_l(x_i) )$
				\State $\mu\gets\mu + \eta_d(1^T\tau - 1 )$
				\EndFor
				\State Output
			\end{algorithmic}
		\end{algorithm}
	\end{minipage}
\end{figure}
{\bf Fast projections when batch size, $B$ is $2$.}
Let us denote by $B$, the minibatch size used to solve \eqref{eq:fin_lag} and consider
the setting where we have $S^+$ and $S^-$ as used in Probs. \eqref{eq:prob1}-\eqref{eq:prob2}.
In this case, we can randomly draw a sample $i$ from $S^+$ and another $i'$ from $S^-$.
Observe that only one coordinate of $\tau$, namely $\tau_i$, changes after the primal update.
Hence, the projection $\Pi_C$ is given by the following simple rule:
after the primal update,
if $\epsilon<0$, we simply set $\epsilon=0$ leaving $\tau_i$ unchanged;
on the other hand if $\epsilon>0$, then we have {\bf two} cases:
\begin{inparaenum}[\bfseries {\text Case} (1)]
\item If $\tau_i\leq \epsilon$, then do nothing, and
\item If $\tau_i\geq \epsilon$, then set $\tau_i=\epsilon=\left(\tau_i+\epsilon\right)/2$. 
\end{inparaenum}
In other words, the projection is extremely easy. 
  
\begin{remark}
  We can indeed use multiple samples (and take the average) to compute gradients of weights $W,\epsilon$, but only
  update one $\tau_i$ with this scheme. 
\end{remark} 

{\bf Generalizing the projection algorithm to larger batch sizes: $B>2$.}
Note that projections onto sets defined by linear inequality constraints can be performed
using standard (convex) quadratic programming solvers. But in the worst case,
such solvers have a time complexity of $O(n^3)$.
It is not practical to use such solvers efficiently in end-to-end training of
widely used neural network architectures. It turns out that a much simpler scheme will work well. 

Define $\uptau:=[\epsilon; \tau]\in\R^{n+1}$, and $\upgamma:=[\delta;\gamma]\in\R^{n+1}$.
The Euclidean projection amounts to solving the following quadratic program:
\begin{align}
  \Pi_{C}(\uptau)=\arg\min_{\uptau} \| \uptau-\upgamma\|_2^2 \quad s.t.\quad \uptau\in C.\label{eq:proj_1}
\end{align}
Our projection algorithm to solve \eqref{eq:proj_1} is given in Algorithm \ref{algorithm2}
which requires {\em only} a sorting oracle.
The following lemma shows the correctness of Algorithm \ref{algorithm2}. 
\begin{lemma}There exists an $O(B\log B)$ algorithm to solve \eqref{eq:proj_1} that requires {\bf 
  only sorting and thresholding} operations.  \label{lemma:proj}
  \begin{proof} Please see appendix.
\end{proof}
\end{lemma}
\begin{figure}[!t]
	\begin{minipage}{\textwidth}
		\begin{algorithm}[H]
			\centering
			\caption{Projection operator: $\Pi_C(\uptau) $}\label{algorithm2}
			\begin{algorithmic}[1]
				\State Input: $\uptau, B>2$
				\State Output: $\Pi_{C}(\uptau)$
				\State If $\uptau_1<0$, set $\uptau_1=0$. 
				\State Compute largest $k^*$ such that
				$\uptau_{k^*+1} \leq \frac{1}{k^*}\sum_{i=1}^{k^*}\uptau_{i}$ by sorting $\{\uptau_i:i>1\}$.
				\State Output: Set $\bar{\mathfrak{o}}= \frac{1}{k^*}\sum_{i=0}^{k^*}\uptau_{i}$,
				\State Return $\Pi_{C}(\uptau_i)$ where
				\begin{subequations}
					\begin{align}
						\Pi_{C}(\uptau_i) &= \bar{\mathfrak{o}} \text{ for } i \leq k^* \text{ or }\\
						\Pi_{C}(\uptau_i) &= \uptau_i \text{ for } i > k^*. 
					\end{align}
				\end{subequations}
			\end{algorithmic}
		\end{algorithm}
	\end{minipage}
\end{figure}
\begin{figure*}[!t]
	\centering
	\begin{subfigure}[!t]{0.22\linewidth}
		\cfbox{black}{\includegraphics[width=1\linewidth, trim = 1cm 1cm 1cm 1cm]{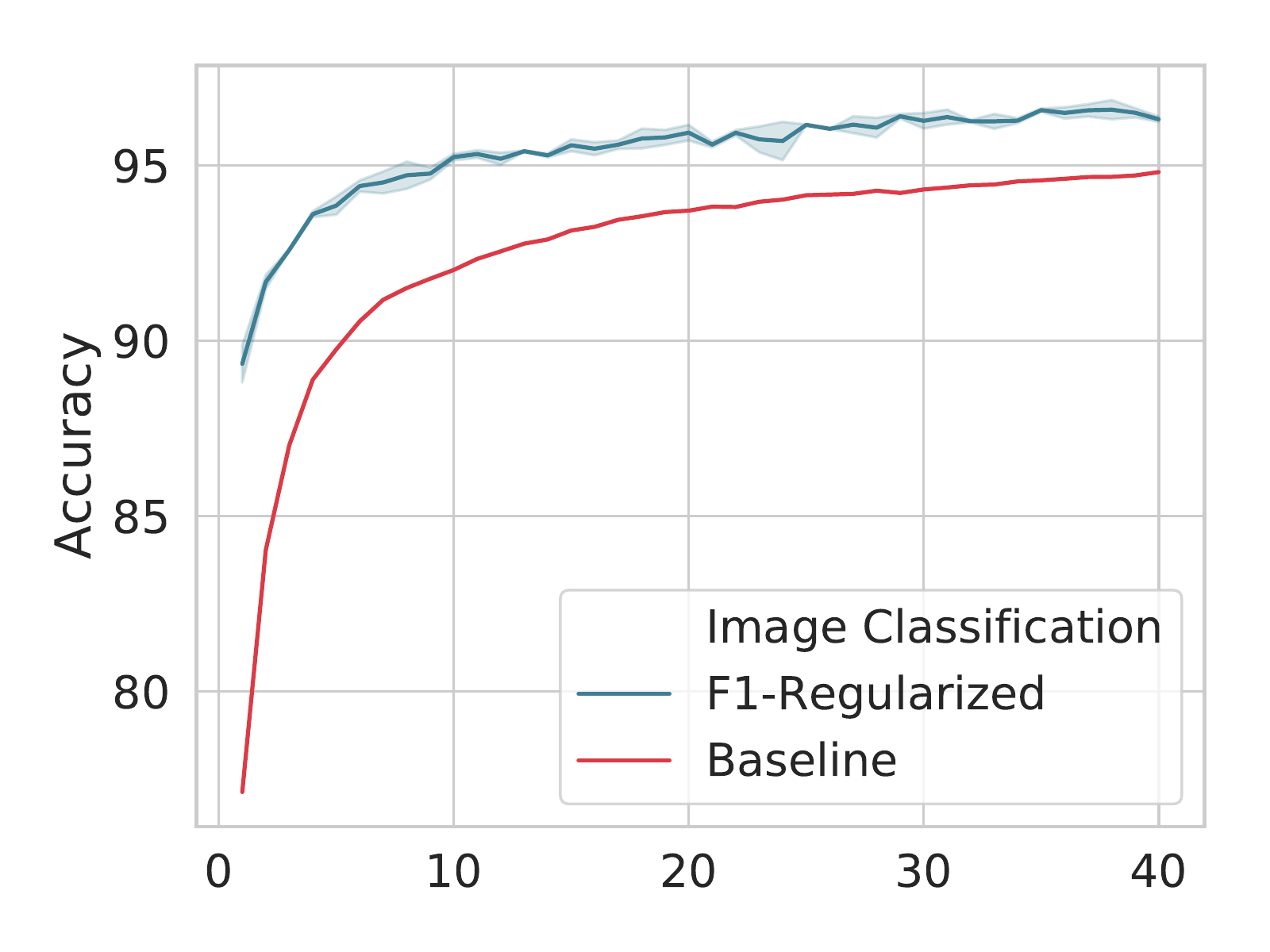}}
	\end{subfigure}		
	\begin{subfigure}[!t]{0.22\linewidth}
		\cfbox{black}{\includegraphics[width=1\linewidth, trim = 1cm 1cm 1cm 1cm]{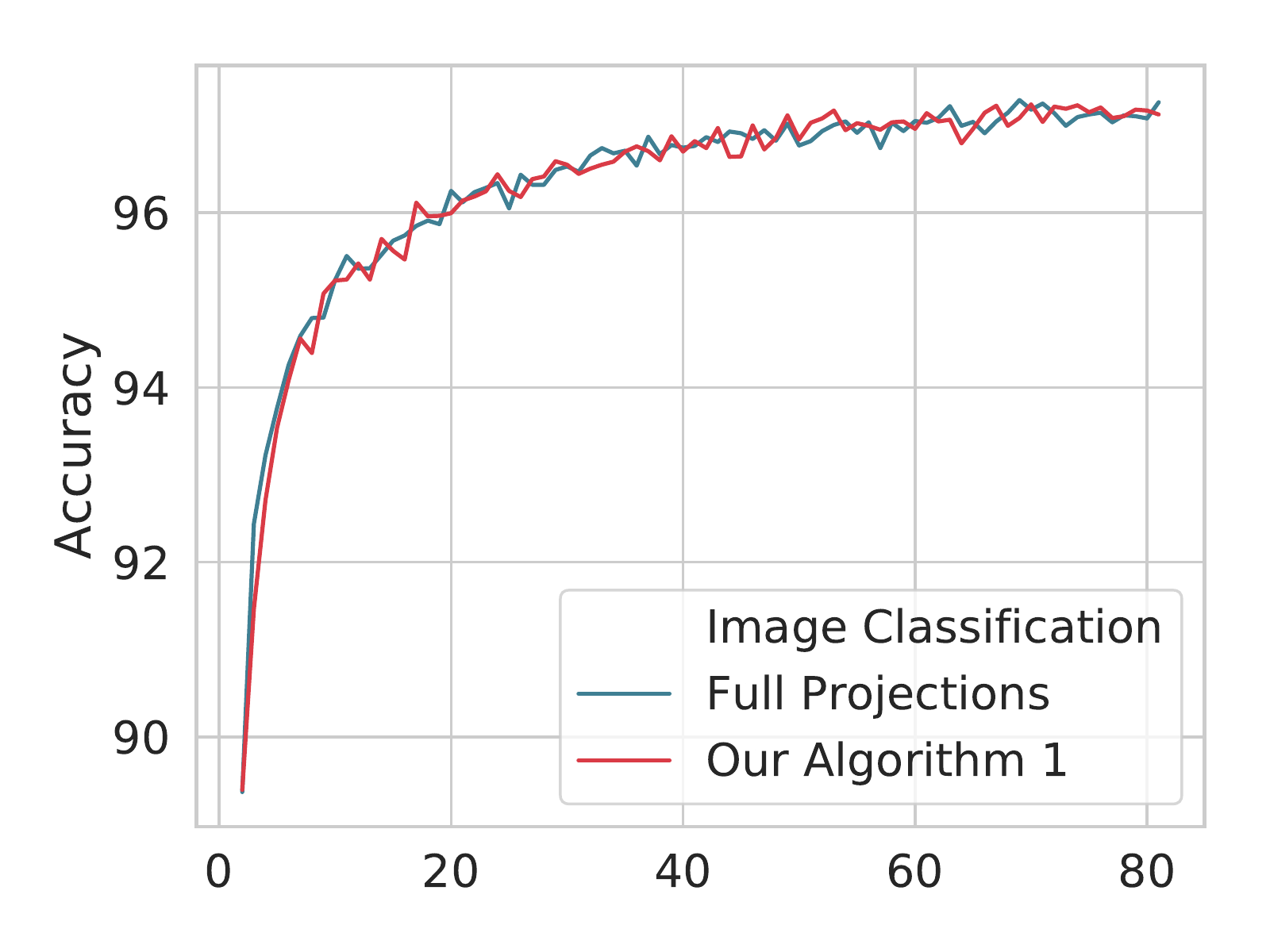}}
	\end{subfigure}		
	\begin{subfigure}[!t]{0.22\linewidth}
		\cfbox{green}{\includegraphics[width=1\linewidth, trim = 1cm 1cm 1cm 1cm]{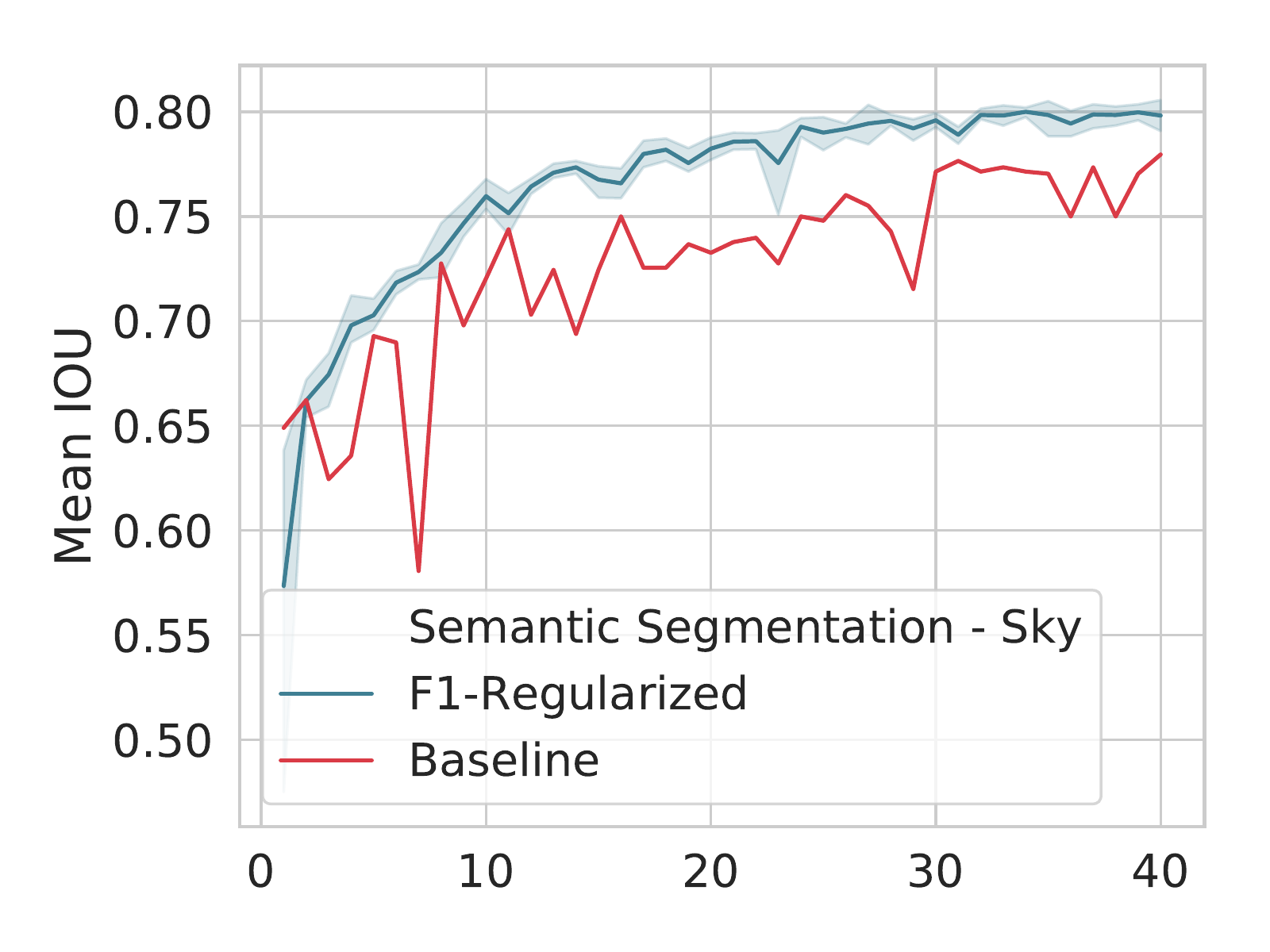}}
	\end{subfigure}	
	\begin{subfigure}[!t]{0.22\linewidth}
		\cfbox{green}{\includegraphics[width=1\linewidth, trim = 1cm 1cm 1cm 1cm]{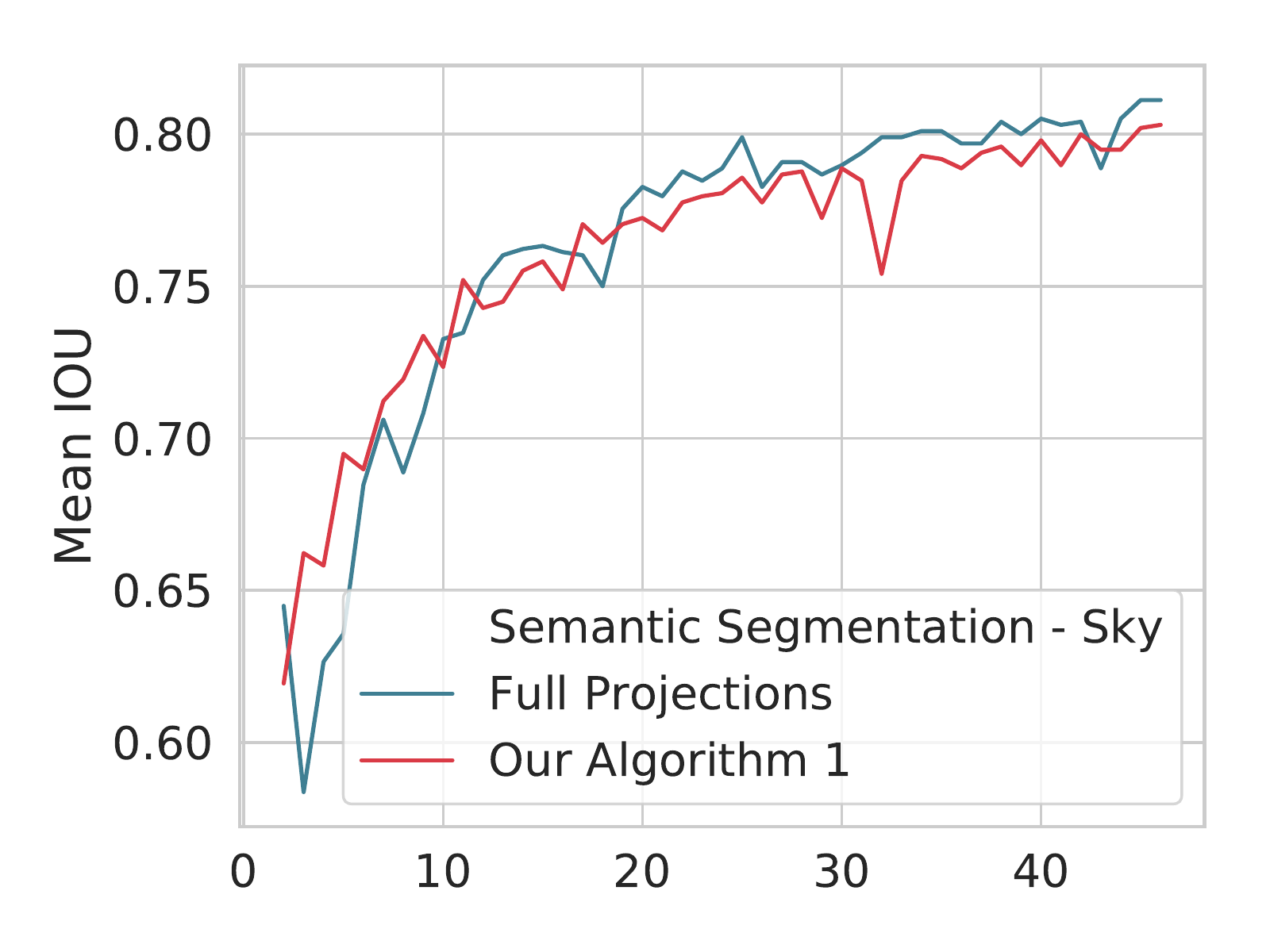}}
	\end{subfigure}		
	\caption{\label{fig:class_sky} (From Left) Performance of our Reparameterization on $\score-$MNIST (Cols. $1\& 2$) and MS-COCO class ``Sky'' (Cols. $3\& 4$) datasets: {$\bf x$-axis denotes the Epoch in all the plots}.  Observe that across both the datasets, Reparameterized Dual Ascent Algorithm (red) \ref{algorithm1}: (i) (Cols. $1\&  3$) obtains high quality solutions than the Baseline (blue) \cite{long2015fully}; (Cols. $2 \& 4$) and requires significantly less projections (once every epoch from Theorem \ref{thm1}).  }
\end{figure*}
\subsection{Obtaining a numerical optimization scheme}
With the results in the preceding
section, we now describe below an efficient approach to solve \eqref{eq:obj_gen}. 

{\bf Reparameterized Dual Ascent.} Using the projection operator above (Alg. \ref{algorithm2}),
our final algorithm to solve $\score-$Measures regularized problem \eqref{eq:obj_gen}
is given in Algorithm \ref{algorithm1} where we use $\eta_d$ to denote the dual step size.
The following theorem characterizes the convergence rate of Algorithm \ref{algorithm1}. 
\begin{theorem} Assume that $\left\|\nabla_W\text{loss}(W;x_i,y_i)\right\|_2\leq G_1 $,
  and $\text{Var}_i\left(\left\|\nabla_W\text{loss}(W;x_i,y_i)\right\|_2\right)\leq \sigma$
  in the ERM term in Prob. \eqref{eq:obj_gen}. Then,
  Alg. \ref{algorithm1} converges to a $\epsilon-$approximate (local) solution of
  Prob. \eqref{eq:obj_gen} in $O(1/\sqrt{T}) $ iterations.\label{thm1}
  \begin{proof}We will use Theorem 1 in \cite{mahdavi2012stochastic} with $\rho=G_2=\sqrt{d}$ and $C_2=d$, see appendix. 
\end{proof}
\end{theorem}

\begin{remark} (Implementation details of Alg. \ref{algorithm2}.)
  In the appendix, based on Lemma \ref{lemma:proj},
  we describe a one pass scheme to compute $k^*$ in Step 4 of Alg. \ref{algorithm2}.
  So, the time complexity of Alg. \ref{algorithm2} is $O(B\log B)$. 
\end{remark}

{\bf Computational Complexity.} Assume that each gradient computation is $O(1)$, Alg. \ref{algorithm1} requires $T=O(1/\epsilon^2)$ iterations to compute
a local $\epsilon$-optimal solution (inner loop). We use a constant number of epochs $E$ (outer loop), so the total cost is $O(E (1/\epsilon^2 + n \log n))$. Please see supplement for more details. 

{\bf Discussion.}
The reader will notice that
the projection step in Algorithm \ref{algorithm1} is {\em outside} the inner loop,
whereas classical Lagrangian based methods guarantee convergence when the projection is performed
in the {\em inner} loop. 
One advantage of having the projection outside the inner \code{for} loop is that
SGD type methods allow for faster optimization of the Lagrangian
with respect to the primal variables $W$.
That is, it is now known that constant step size
policies guarantee faster convergence of the inner \code{for} loop
under some structural assumptions \cite{dieuleveut2017bridging}.
In any case, we give a detailed description of our Algorithm \ref{algorithm1} in the
supplement with a discussion of the trade-off between
using projections in the inner \code{for} loop versus the outer \code{for} loop in Algorithm \ref{algorithm1}. 

\subsection{Approximating $\Pi_C$ for Backpropagation}
Recall that our primary goal is to solve our optimization problem \eqref{eq:fin_lag} using backpropagation based methods. In the previous sections, we argued that SGD can be used to solve \eqref{eq:fin_lag}. In particular, we showed in Theorem \ref{thm1} that Algorithm \ref{algorithm1} requires fewer projections compared to the classical PGD algorithm. It turns out that the simplicity of the projection operator $\Pi_C$ can be exploited to avoid the double \code{for} loops in Algorithm \ref{algorithm1}, thus enabling backpropagation based training algorithms to be effectively deployed. Please see supplement for all the details.
	\section{Experimental Evaluations}
\label{sec:exps}
{\bf Overview.} We show experiments on three different tasks:
$\score-$Measures defined using the $F_1$ metric can be efficiently optimized using our
Reparameterized Dual Ascent Algorithm \ref{algorithm1} in large scale vision problems while
strongly improving the empirical performance.
The {\bf first} set of experiments is designed to show that the
$F_1$ metric based objectives
may benefit existing classification models using convex classifiers.
The {\bf second} set of experiments further evaluates the performance of
Alg. \ref{algorithm1} for nonconvex models.
Our goal here is to show that nondecomposable regularizers {\em can}
be used to stabilize models without sacrificing on the overall accuracy.
In the {\bf third} set of experiments, we show that architectures
used for Semantic Segmentation can be trained using Alg. \ref{algorithm1}.
In all our experiments, we used ADAM optimizer (with default parameters)
to train with primal step sizes of $\eta_{\tau} =\eta_{\epsilon}=\eta_{W}=10^{-2}$, and dual step sizes of $\eta_\lambda= 10^{-3}$, and $\eta_\mu= 10^{-5}$.
We report the  results for the regularization parameter taking values
$\alpha= \{10^{-2},10^{-3},10^{-4}\}$. Baseline corresponds to $\alpha=0$.

	\begin{figure*}[!tb]
	\centering
	\begin{subfigure}{\textwidth}
		\smallskip
		\begin{subfigure}[t]{0.24\linewidth}
			\centering
			\includegraphics[width=\linewidth]{images/triplets/000000053626.jpg}
		\end{subfigure}
		\begin{subfigure}[t]{0.24\linewidth}
			\centering
			\includegraphics[width=\linewidth]{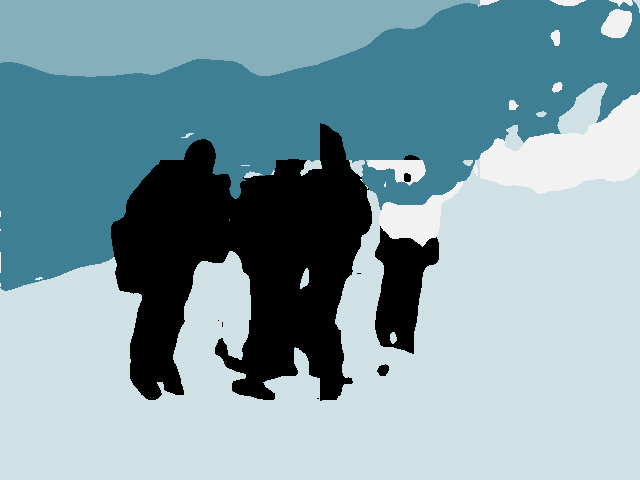}
		\end{subfigure}
		\begin{subfigure}[t]{0.24\linewidth}
			\centering
			\includegraphics[width=\linewidth]{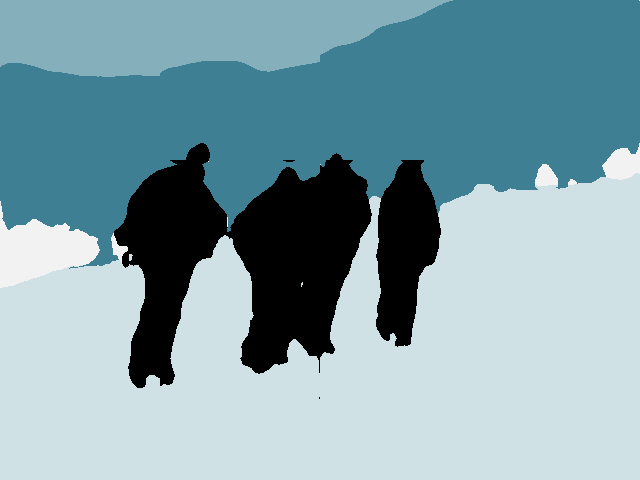}
		\end{subfigure}
		\begin{subfigure}[t]{0.24\linewidth}
			\centering
			\includegraphics[width=\linewidth]{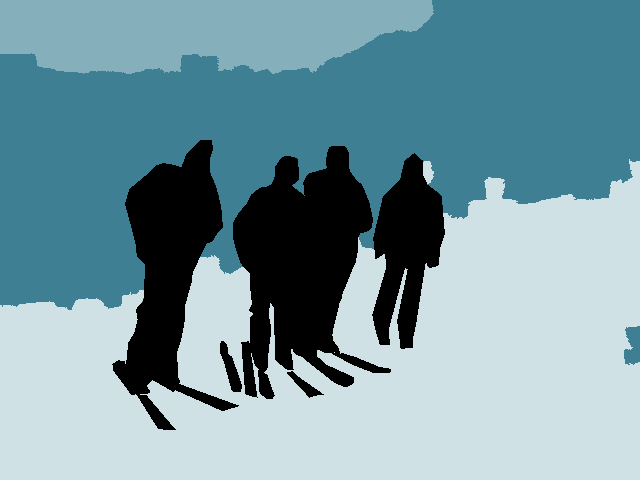}
		\end{subfigure}\\
		\begin{subfigure}[t]{0.24\linewidth}
			\centering
			\includegraphics[width=\linewidth]{images/triplets/000000359781.jpg}
		\end{subfigure}
		\begin{subfigure}[t]{0.24\linewidth}
			\centering
			\includegraphics[width=\linewidth]{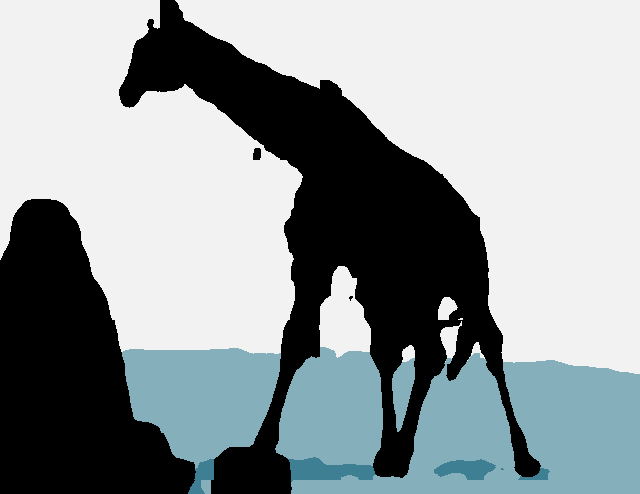}
		\end{subfigure}
		\begin{subfigure}[t]{0.24\linewidth}
			\centering
			\includegraphics[width=\linewidth]{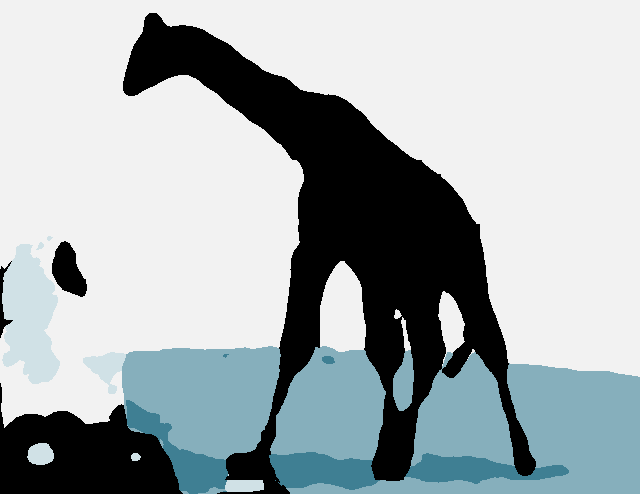}
		\end{subfigure}
		\begin{subfigure}[t]{0.24\linewidth}
			\centering
			\includegraphics[width=\linewidth]{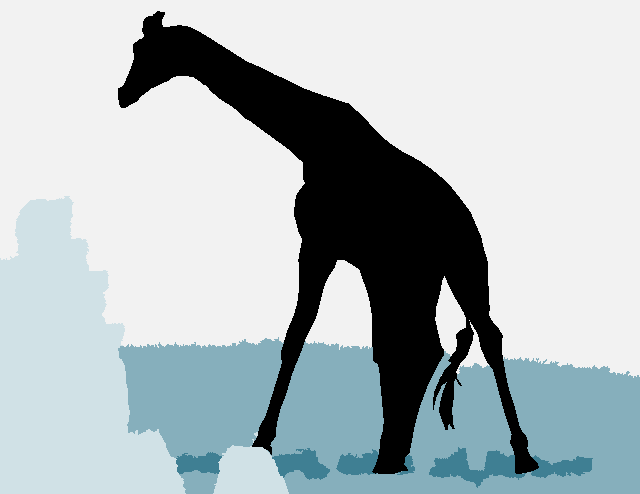}
		\end{subfigure}
	\end{subfigure}\\
	\caption{\label{fig:mscoco-name} \color{black}  Qualitative results on MSCOCO stuff segmentation benchmark.  \label{fig:sub-default} \color{black} Colors (except black) indicate different ``stuff'' categories in an image. From left to right: original image, baseline,
		our result and ground truth.}
\end{figure*}
\paragraph{Experiment 1) Improving Linear Classifiers using $\score-$Measures\label{exps:smnist}}
{\bf Dataset.} Consider  
a dataset with distinct classes: MNIST with 10 classes (digits).
In MNIST, the {\em size} of the foreground (pixels with nonzero intensities)
is about the same for all images (and each class) in MNIST --  the distribution of foreground size $s'$ is
{\em unimodal} (see appendix). To see the benefits of $\score-$measures,
we create an augmented dataset called $\score-$MNIST.
We added a set of images in which the foreground is double the original
size in MNIST simply by rescaling the foreground to get a {\bf two mode} $s'$.
 
{\bf How to obtain size s?}  To get $s\in\{0,1\}$, we can simply threshold $s'$ using the empirical mean.

{\bf Model and Training.} We used a single layer neural network $f$ to predict $y$ from $x$ but
added an extra node to account for the $\score-$Measure term in \eqref{eq:obj_gen}.
That is, our total loss is the sum of the softmax cross-entropy ERM loss and the $\score-$Measure (using $F_1$ metric).
We trained using Alg. \ref{algorithm1} for $40$ epochs with a minibatch size of $100$.  

{\bf Results.} We see the benefits provided by the $F_1$ metric as a data-dependent
regularizer in Fig. \ref{fig:class_sky}.
Column 1 shows that our model is uniformly
more accurate than the baseline throughout the training process.
Moreover, Column 2 compares Alg. \ref{algorithm1} with classical dual ascent procedures from \cite{cotter2018optimization}.
Here, full projections refers to computing $\Pi_C(\cdot)$ after every inner iteration in Alg. \ref{algorithm1}.
Clearly, we can see that Algorithm \ref{algorithm1} obtains high quality solutions but needs {\bf one projection operation} every epoch.

{\bf Takeaway \#1.} Models obtained using Alg. \ref{algorithm1} are more accurate and stable for linear classifiers.

	\paragraph{Experiment 2) Improving one class Segmentation using $\score-$Measures.}  We consider the task of the pixel-wise contextual labeling of an image. 
We found that the ``sky'' category in the {\em MSCOCO stuff} (2017)  dataset \cite{caesar2018cvpr} has a high variance over the samples in terms of size: so,
taking this property into account seems sensible. 
So, we use {\bf only} the images in the ``sky'' category \cite{caesar2018cvpr}.  

{\bf How to obtain size $s$?} We first computed the empirical distribution of number of pixels $s'$ that are labeled as sky in the training set.
We then pick the mean of $s'$ to be the threshold to obtain a binary $s$.

{\bf Model and Training.}
We utilize SegNet \cite{badrinarayanan2017segnet,mshahsemseg}, a deep encoder-decoder architecture for semantic segmentation for its simple design and ease of use. 
As before, we add a fully-connected layer at the end of the model to incorporate the $\score-$Measure specified by  $F_1$ metric.
We trained our models for 40 epochs with a learning rate of 0.01 and batch size 7. 

{\bf Results.}
Figure \ref{fig:class_sky} Column 3 shows our results averaged over the thresholds $\{0.55,0.6,0.65\}$. We can see that our models uniformly outperforms the baseline with a $80\%$ mean IOU (over $77.6\%$ baseline). Furthermore, observe that our Alg. \ref{algorithm1} is much more stable than the baseline throughout the training process even in nonconvex settings
while requiring the fewest projection steps. 

{\bf Takeaway \#2.}
Performance boosts from the F1-regularizer carries over from the simple convex classification task to the pixel-wise classification task
with a deep neural network.

\paragraph{Experiment 3) Improving Semantic Segmentation with Nondecomposable Regularizers} 

{\bf Dataset.} We are now ready to consider semantic segmentation with multiple classes in each image. 
Our dataset consists of $164K$ images that belong to any of the ``stuff'' classes in the MSCOCO ``Stuff'' dataset \cite{caesar2018cvpr}. We downsampled the training images  to $106\times 160$ size
to reduce training time. 

{\bf How to obtain size $s$?}
The {\em volume of stuff $c$} denoted by $s'^c$ is measured by the number of pixels that belong to $c$.
We observed that $s'^c$  is close to $0\%$ for most $c$ (see appendix). So,
we picked a threshold of $\approx 0.05$ to obtain a binary $s^c\in\{0,1\}$ for each class $c$.
Then, we use the majority vote provided by all classes present in an image to obtain $s\in\{0,1\}$ for individual images --
corresponding to ``big/small''.
That is, if the majority of the classes present inside the image are ``big'' (as determined by the threshold $s=0.05$), then we assigned the image to be ``big'' and vice-versa.

{\bf Model and Training.}
We use DeepLabV3+  model proposed in \cite{chen2018encoder} for training. DeepLabV3+ is a popular model for semantic segmentation (needs no CRF post-processing)
and can be trained end-to-end. We used a minibatch size of $144$
for baseline and $120$ for our $F_1$ regularized models. 

{\bf Results.} Figure \ref{fig:mscoco-name} Column 1 shows the quantitative results of our experiment averaged over the thresholds $\{0.01,0.03,0.05,0.07,0.09\}$.  Columns 2-4 shows some of our qualitative results.  See appendix for quantitative results on MSCOCO stuff segmentation benchmark. {\bf Our mean IOU improves upon the state of the art
reported in \cite{chen2018encoder} by $\approx 10\%$ on MSCOCO 164K while being stable.}

{\bf Takeaway \#3.} On the MS COCO 164K Stuff dataset, we achieve state of the art results with $0.32$ Mean IOU (vs. $0.30$ current {\color{black}state of the art mean IOU  in \cite{chen2018encoder}}).

	\section{Conclusions}

While nondecomposable data-dependent regularizers are variously beneficial and needed in a number of
applications, their benefits cannot often be leveraged in large scale settings due to
computational challenges. Further, the literature provides little guidance on mechanisms
to utilize such regularizers within the deep neural network architectures
that are commonly used in the community. In this paper,
we showed how various nondecomposable regularizers may indeed permit
highly efficient optimization schemes that can also directly
benefit from the optimization routines implemented in mature software libraries used in
vision and machine learning. We provide a technical analysis of the algorithm and show that
the procedure yields state of the art performance for a semantic segmentation task, with
only minimal changes in the optimization routine.

	\bibliography{references}
\bibliographystyle{natbib}
	
{\section*{Appendix}}
\section{Extending our results to other Metrics}
{\em Point Metrics.} To see how to extend our results to say R@P$_{\alpha}$ metric, we simply note that both recall and precision can be represented as {\em rate} constraints, as discussed in \cite{cotter2018optimization}. This means that we can simply reparameterize these terms as shown in the main paper, and does not require a linear fractional transformation that we need to use for $F_{\beta}$ type metrics.  

{\em Integral Metrics.} To extend to AUCROC type metrics that are defined as integrals over the point metrics, we can use Nystrom Approximation. That is, we first choose a sufficiently fine grid (at many $\alpha$'s), and then reparameterize the R@P at each point ($\alpha$) in the grid. Hence, the convergence guarantee in Theorem 1 apply in both of these settings.

\section{Missing details from Section 3 in the Main Paper}
\subsection{Generalizing to finite $|\score|=p>2$}
First we discuss on how to generalize our formulation when finite $|\score|=p$. For our purposes, we may assume from now on that $f$ is linear, parameterized by $W$ (without of loss generality) since we can use the reparameterization shown in the main paper. Furthermore as in section 3 of the main paper, we may ignore the ERM term. When $p>2$, there are several ways to generalize the $F_1$ metric, see Section 6 \cite{lipton2014optimal}. Essentially, the $F_1$ score may simply be between each pair of classes in the training data. This will lead to constraints of the form,\begin{align}
\tau_i\leq W_j\cdot (x_i); \,\forall~i\in  S^j,j=1,\dots,p, \label{eq:multi}
\end{align}
where $W_j$ represents the $j-$th column of $W$ ($j-$th size predictor) and $W_j\cdot (x_i)$ represents the margin of example $i$ with respect to the size $j$ predictor given by $W_j$. Following the reparameterization technique in Section 3 in the main paper, we simply dualize the complicating constraints shown in \eqref{eq:multi} by introducing dual variables $\lambda_{ij}$.
\begin{remark} Observe that Algorithm 1 remains unchanged after the addition of dual variables. Naturally, the convergence of Algorithm 1 depends on the number $p$. Without further assumptions on the data (such as separability) it is currently not possible to obtain convergence rate with no dependence on $p$, see \cite{allen2018natasha}.
\end{remark}
\subsection{Proof of Lemma 1 in the main paper} Recall that the Euclidean projection amounts to solving the following quadratic program:
\begin{align}
\Pi_{C}(\uptau)=\arg\min_{\uptau} \| \uptau-\upgamma\|_2^2 \quad s.t.\quad \uptau\in C.\label{eq:proj_1}
\end{align}
\begin{lemma}There exists an $O(B\log B)$ algorithm to solve \eqref{eq:proj_1} that requires {\bf 
		only sorting and thresholding} operations.  \label{lemma:proj}
	\begin{proof}
		WLOG, we can assume that
		$ \upgamma_2\geq \upgamma_3\geq\cdots\geq \upgamma_{n+1}$ since when
		given any $\gamma$, we may rearrange the coordinates.
		We may also assume that for any $i>1$, $\upgamma_{i}\neq \upgamma_{i+1}$
		by picking an arbitrary order for
		coordinates with the same value. 
		Denoting the lagrange multipliers for the constraints $C$ in in \eqref{eq:proj_1}, the feasibility conditions for \eqref{eq:proj_1} can be written as,
		\begin{align}
		&       \text{{\bf Primal.}} \quad \uptau_1 \geq \uptau_i ~\forall~i, \uptau_1\geq 0,
		&    \text{\bf Dual.} \quad \dual \geq 0.  \label{eq:proj_d_1}
		\end{align}
		The first order necessary conditions can be written as,   \begin{align}
		\uptau_i-\upgamma_i+\dual_i = 0~\forall~i,~~	\uptau_1-\upgamma_1-\dual_1 -\sum_{i=2}^{n+1}\dual_{i} =0.\label{eq:proj_d_2}
		\end{align}Finally, the complementarity slackness condition between the primal and dual variables can be written as,\begin{align}
		\dual_i(\uptau_i-\uptau_1) = 0~\forall~i~~	\dual_1 \uptau_1 = 0.\label{eq:proj_d_3}
		\end{align}
		From \eqref{eq:proj_d_2}, we have that $\dual_{i} = \upgamma_i-\uptau_i$
		which implies that 
		and $\uptau_1=\upgamma_1+\sum_{i=1}^{n+1}\dual_{i}\geq \upgamma_1 $.
		From \eqref{eq:proj_d_3}, either $\dual_{i}=0$ or $\uptau_i=\uptau_1$, and $\dual_1 =0$.
		Let $I:=\{i\neq 1:\uptau_i=\uptau_1 \}$ be the (active) set of indices at the optimal solution.
		Given $I$, we can easily find $\uptau_1$ by solving the following 1D quadratic problem
		\begin{align}
		\min_{\uptau_1\geq 0} \sum_{i\in I\cup \{1\}} \frac{1}{2}(\uptau_1-\upgamma_i)^2  \implies \uptau_1 = \frac{1}{|I|+1} \sum_{i \in I\cup \{1\}}\upgamma_i.
		\end{align} 
		So, it is now
		clear that in order to satisfy the feasibility conditions \eqref{eq:proj_d_1},
		we simply choose the first $k+1$ coordinates of $\upgamma_1$ to be the index set $I$, and
		{set $\uptau_1$
			to be the average of $\{\upgamma_i\}$ where $i\in I$.}
	\end{proof}
\end{lemma}
\subsection{Algorithm 1: Implementation and Computational Complexity}
Now, we will provide the full details of Algorithm 1 in the main paper here in Algorithm \ref{algorithm1}. We now state the definition of our lagrangian (for convenience), \begin{align}
\max_{\lambda\geq 0,\mu}\min_{(\tau,\epsilon)\in C,W} L(\tau,w,\epsilon,\lambda,\mu).\label{eq:fin_lag}
\end{align}
\begin{figure}[!t]
	\begin{minipage}{\textwidth}
		\begin{algorithm}[H]
			\centering
			\caption{Reparameterized Dual Ascent for solving  \eqref{eq:fin_lag}}\label{algorithm1}
			\footnotesize
			\begin{algorithmic}[1]
				\State Input: $X,Y,S,$ trainable parameters: $W,\uptau$                   
				\State Initialize variables $W,\uptau,T,\text{Epochs},\eta_d$
				\For{$e= 1,\dots, \text{Epochs}$}
				\For{$t= 1, \dots, T$}
				\State Choose $x_j, j\in S^-$ with probability $p_j=B/2|S^-|$ and compute (i) $g_j^w=\nabla_w\max(0,\epsilon+w\cdot x_j)$, (ii) $g_j^{\epsilon} =\nabla_\epsilon \max(0,\epsilon+w\cdot x_j)$.
				\State Choose $x_i, i\in S^+$ with probability $p_i=B/2n$ and compute (i) $g_i^w=-\lambda_ix_i$, (ii) $g_i^{\tau_i} =\mu +\lambda_i$.
				\State ``Full'' Projections requires to project every iteration, $\uptau \gets \Pi_{C}(\uptau)$ using Algorithm \ref{algorithm2}
				\EndFor
				\State Our algorithm requires only this: $\uptau \gets \Pi_{C}(\uptau)$ using Algorithm \ref{algorithm2}
				\State $\lambda\gets \lambda + \eta_d(\tau - w_{l+1}\cdot a_l(x_i) )$
				\State $\mu\gets\mu + \eta_d(1^T\tau - 1 )$
				\EndFor
				\State Output
			\end{algorithmic}
		\end{algorithm}
	\end{minipage}
\end{figure}
{\em Computational Trade-off.} We can now clearly see the trade-off between doing projections done at every iteration (referred to as ``Full'' projections) and our reparameterized dual ascent procedure that requires only one projection for one epoch. As we can see from Algorithm \ref{algorithm2} here, the computational burden of projection is due to sorting that is required. if $B=\log n$, then doing full projections is attractive, whereas for large batch training, say $B=\sqrt{n}$, our algorithm \ref{algorithm1} -- reparameterized dual ascent is a much more efficient strategy. 

{\em Counting Projections.} Keeping all other parameters the same, the computational complexity of full projections method will be $O(TB\log B)$ whereas it is only $O(n\log n)$ in the worst case for every epoch. This means that, if we are closer to a local minima (small $T$), then using full projections is cheaper since it is independent of the dataset size $n$. However, as mentioned in the main paper $T=O(1/\epsilon^2)$ in the worst case to obtain an $\epsilon-$approximate (gradient norm) solution of the lagrangian $L$ (inner loop). Hence, Algorithm 1 is preferable if we are far from the optimal solution (random initialization of weights) and use full projections when using a pretrained model to fine tune the weights.

\subsection{Constants in the Proof of Theorem 2} We recall the theorem statement.
\begin{theorem} Assume that $\left\|\nabla_W\text{loss}(W;x_i,y_i)\right\|_2\leq G_1 $,
	and $\text{Var}_i\left(\left\|\nabla_W\text{loss}(W;x_i,y_i)\right\|_2\right)\leq \sigma$
	in the ERM term in Prob. \eqref{eq:obj_gen}. Then,
	Alg. \ref{algorithm1} converges to a $\epsilon-$approximate (local) solution of
	Prob. \eqref{eq:fin_lag} in $O(1/\sqrt{T}) $ iterations.\label{thm1}
	\begin{proof}We will use Theorem 1 in \cite{mahdavi2012stochastic} with $\rho=G_2=\sqrt{d}$ and $C_2=d$. 

With a slight abuse of notation we will use the matrix $C$ to represent the feasible set $C$ since there is no confusion here. Hence the feasible set can be written as,\begin{align}
\uptau\in \R^{n+1} \quad \text{s.t.} \quad \overbrace{\begin{bmatrix}
-1       & 1 & 0 & &\dots & 0 \\
-1       & 0 & -1 & 0&\dots & 0 \\
\vdots &\ddots&&&&\vdots\\
-1      & 0 & \cdots & & &  0
\end{bmatrix}}^{C\in\R^{n+1\times n+1}}
\begin{bmatrix}
\uptau_1\\
\uptau_2 \\
\vdots \\
\uptau_{n+1}
\end{bmatrix}\leq 0.
\end{align}
with \begin{align}
\left\| \nabla_{\uptau} (C\uptau)\right\|_2 = \|C \|_2 \sim \sqrt{d}.
\end{align} Since linear functions are $1-$lipschitz, we have the required constants for Theorem 1 in \cite{mahdavi2012stochastic} for our choice of $C$ to bne constants as $\rho=G_2=\sqrt{d}$ and $C_2=d$.
	\end{proof}
\end{theorem}

\subsection{One Pass Algorithm to compute $\mathfrak{o}$ in Algorithm 2 in the Main Paper}
We will use the following notation to slice coordinates: for a vector $v\in\R^n, i\leq j$, $v[i:j]$ selects the subspace parameterized by coordinates from $i$ to (and including) $j$. We will now provide the sequential algorithm to implement Algorithm 2 in the main paper in here in Algorithm \ref{algorithm2}. Observe that the {\em online} nature of the algorithm suggests that we may simply compute $\mathfrak{o}$ using sampling/sketching procedures, that is, pick a sample of $\{\tau_i\}$ of size $n'$, sort it and compute the threshold. For certain distributions of $\tau$, $\mathfrak{o}$ can be {\em approximated} using $n'$ samples -- independent of the training dataset size $n$ \cite{tropp2015introduction}. 
\begin{figure}	[!tb]
	\begin{minipage}{\textwidth}
		\begin{algorithm}[H]
			\centering
			\caption{Projection operator: $\Pi_C(\uptau) $}\label{algorithm2}
			\footnotesize
			\begin{algorithmic}[1]
				\State Input: $\uptau, B>2$
				\State Output: $\Pi_{C}(\uptau)$
				\State If $\uptau_1<0$, set $\uptau_1=0$. 
				\State Sort $ \uptau[2:n+1]$ in descending order
				\State $\uptau_1\gets \max(\upgamma_1,0)$
				\For{{$t= 1, \dots, n$}}
					\If {$\upgamma_{t+1}\leq \max(\uptau_1,0)$}
						\State $k^*\gets t$ and {\bf Break}						
					\Else
						\State $\uptau_1=\frac{t}{t+1}\uptau_1 + \frac{1}{t+1} \upgamma_{t} $
					\EndIf					
				\EndFor
				\State $\uptau_1\gets \max(\uptau_1,0)$
				\State $\uptau[2:k^*-1] \gets \uptau_1$
				\State $\uptau[k^*:n+1] \gets \upgamma[k^*:n+1]$,
				\State Return $\uptau$
			\end{algorithmic}
		\end{algorithm}		
	\end{minipage}
\end{figure}
\subsection{Approximating $\Pi_C$ for Backpropagation}
Recall that our primary goal is to solve our optimization problem \eqref{eq:fin_lag} using backpropagation based methods. In the previous sections, we argued that SGD can be used to solve \eqref{eq:fin_lag}. In particular, we showed in Theorem \ref{thm1} that Algorithm \ref{algorithm1} requires fewer projections compared to the classical PGD algorithm. It turns out that the simplicity of the projection operator $\Pi_C$ can be exploited to avoid the double \code{for} loops in Algorithm \ref{algorithm1}, thus enabling backpropagation based training algorithms to be effectively deployed. We rely on the following simple, yet crucial observation,\begin{obsn}\label{obsn:relu}Given $\bar{\mathfrak{o}}$  in Algorithm \ref{algorithm2}, $\Pi_C$ can be computed using the rectified linear unit (relu) function as follows,\begin{align}
	\Pi_C(\tau) = \min(\bar{\mathfrak{o}},\tau)=-\max(-\bar{\mathfrak{o}},-\tau)\label{eq:obsrelu}
	\end{align}
	where the $\min$ and $\max$ are applied elementwise.
\end{obsn}
To verify equation \eqref{eq:obsrelu}, note that by definition of $\bar{\mathfrak{o}}$, the first $k^*$ coordinates are no larger than $\bar{\mathfrak{o}}$ and the remaining coordinates are strictly smaller than $\bar{\mathfrak{o}}$. Now we will show how to compute $\bar{\mathfrak{o}}$ using differentiable techniques. Our approach involves four steps: \begin{enumerate}
	\item	{\em (Differentiable Sort.)} The first key step is to use recently developed algorithms to produce an approximate sorting of $\tau$ using Optimal Transport theory, see \cite{cuturi2019differentiable}. In particular, we first sort the input $\tau$ to $\Pi_C$ using the so-called Sinkhorn iteration. From now on, we assume that $\tau$ is sorted in decreasing order of magnitues as required by Algorithm \ref{algorithm2} in Step 4.
	\item {\em (Vector of top $i-$averages.)} Secondly, we compute the vector $\bar{\tau}\in\R^n$ of top $i\in\{1,2,\dots,n\}$ averages of $\tau$. 
	\item {\em (Computing $k^*$ in Algorithm \ref{algorithm2}.)} Define $\tau^s$ to be the vector shifted by one coordinate, that is, the $i-$th coordinate of $\tau^s$ is given by the $i+1-$th coordinate of $\tau$ with a $0$ in the last coordinate. Note that by the monotonicity of $\tau$, we have that,  $\hat{\tau}_i\geq \tau^s_i~\forall~i\leq k^*$.  Hence, we have that $k^*=1^T\hat{\tau}$ where $\hat{\tau}$ is defined as, \begin{align}
	\hat{\tau}:= \max\left(\max\left(\tau^s-\bar{\tau},0\right),1\right).\label{eq:kstar}
	\end{align}
	In words, the inner $\max$ operation compares the top $i$ averages with the $i+1-$th coordinate and sets the bottom $n-k^*$ coordinates to zero whereas the outer $\max$ operation sets the top $k^*$ to one.
	\item {\em (Recovering $\bar{\mathfrak{o}}$.)} Finally, we can easily compute $\bar{\mathfrak{o}}$ as,\begin{align}
	\bar{\mathfrak{o}}=\frac{1^T\text{diag}(\hat{\tau})\tau}{1^T\text{diag}(\hat{\tau})},
	\end{align}
	where $1\in\R^n$ corresponds to the vector of all ones and $\text{diag}(x)$ represents the diagonal matrix with entries given by the vector $x$.
\end{enumerate} Note that computing $\bar{\tau}$ just involves a {\em single} matrix-vector product. That is, given $\tau$, we can compute $\bar{\tau}$ by applying the lower triangular matrix with all entries in the $i-$th row given by $1/i$ to $\tau$. Hence, all the four steps mentioned above can be unrolled into a ``forward pass'', we can compute the gradient with respect to the trainable parameters  of our network using the adjoint method or backpropagation, see \cite{baydin2018automatic}.

\section{Bimodal $\score-$MNIST using Unimodal MNIST} 
\begin{figure*}[!t]
	\centering
	\begin{subfigure}[!t]{0.19\linewidth}
		{\includegraphics[width=1\linewidth]{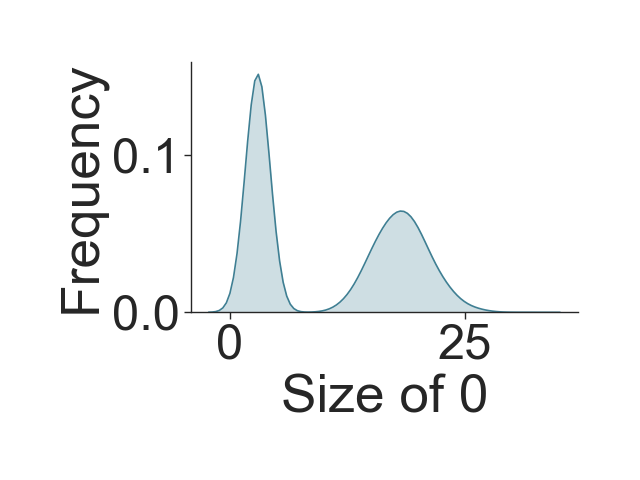}}
	\end{subfigure}
	\begin{subfigure}[!t]{0.19\linewidth}
		{\includegraphics[width=1\linewidth]{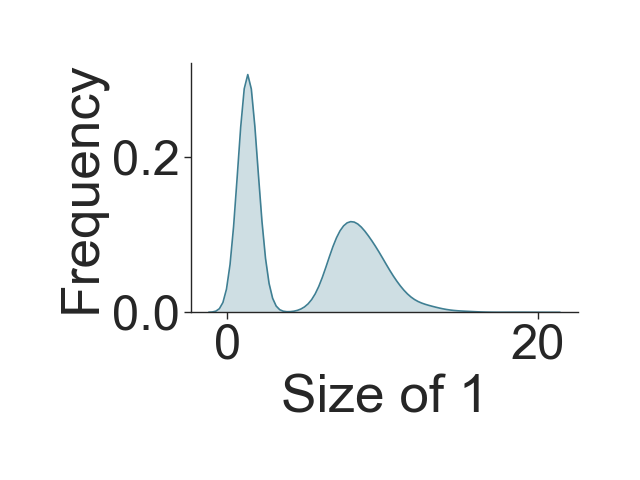}}
	\end{subfigure}
	\begin{subfigure}[!t]{0.19\linewidth}
		{\includegraphics[width=1\linewidth]{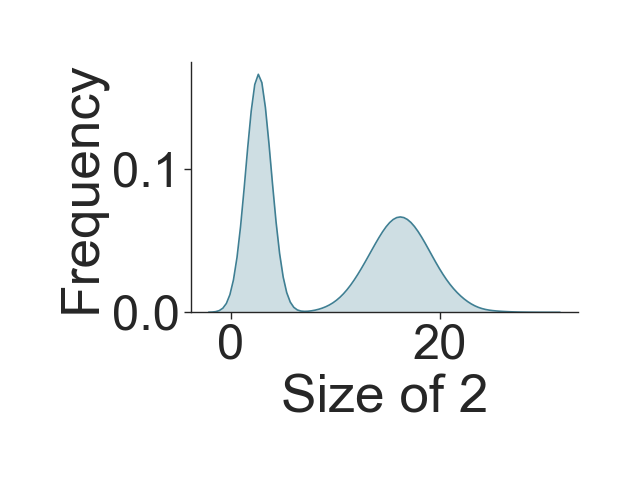}}
	\end{subfigure}
	\begin{subfigure}[!t]{0.19\linewidth}
		{\includegraphics[width=1\linewidth]{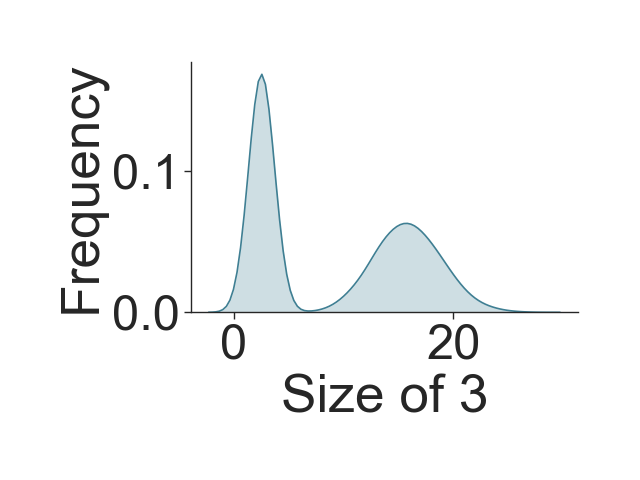}}
	\end{subfigure}		
	\begin{subfigure}[!t]{0.19\linewidth}
		{\includegraphics[width=1\linewidth]{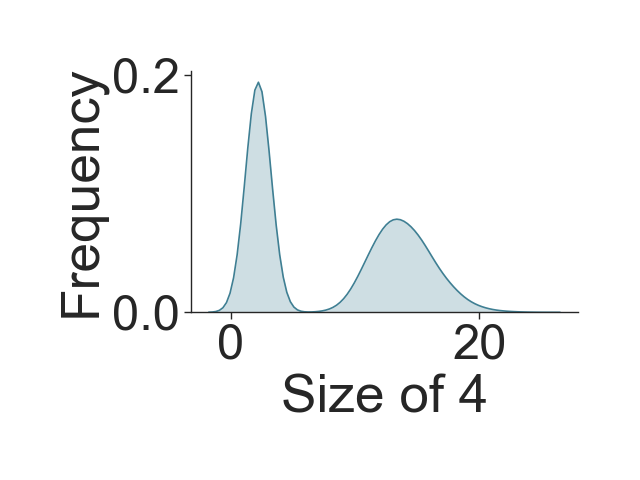}}
	\end{subfigure}
	\begin{subfigure}[!t]{0.19\linewidth}
		{\includegraphics[width=1\linewidth]{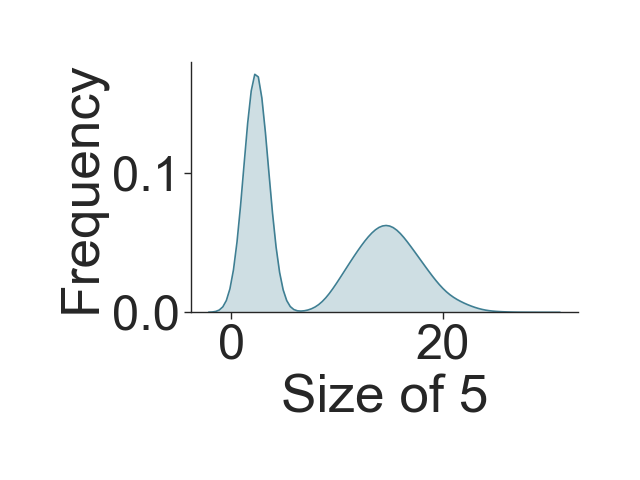}}
	\end{subfigure}	
	\begin{subfigure}[!t]{0.19\linewidth}
		{\includegraphics[width=1\linewidth]{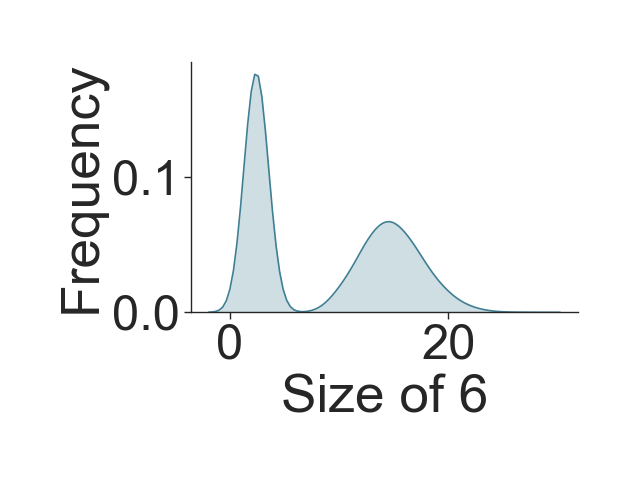}}
	\end{subfigure}
	\begin{subfigure}[!t]{0.19\linewidth}
		{\includegraphics[width=1\linewidth]{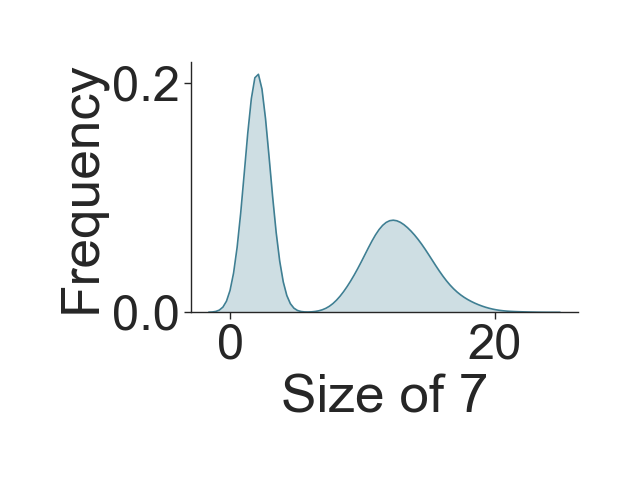}}
	\end{subfigure}	
	\begin{subfigure}[!t]{0.19\linewidth}
		{\includegraphics[width=1\linewidth]{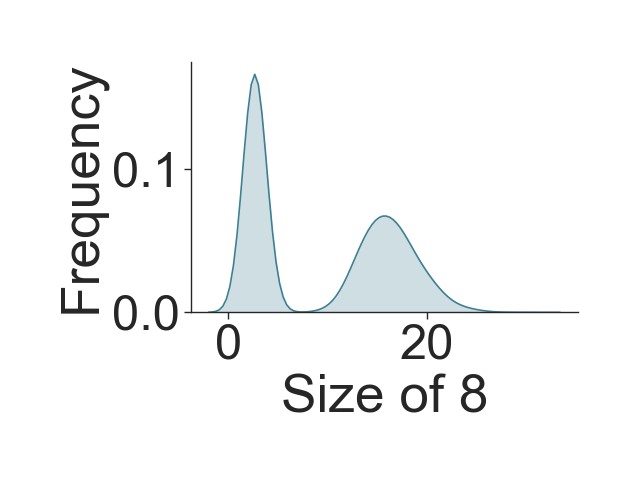}}
	\end{subfigure}
	\begin{subfigure}[!t]{0.19\linewidth}
		{\includegraphics[width=1\linewidth]{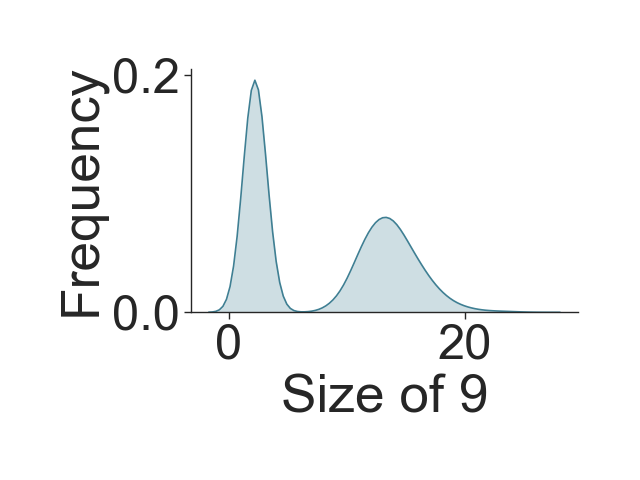}}
	\end{subfigure}	\caption{$\score-$MNIST Histograms.}\label{fig:mnist}
\end{figure*}

Here, we show the histograms of the size of digits in our $\score-$MNIST created using MNIST as discussed in Section 4.1 in the main paper in Figure \ref{fig:mnist}. Observe that while the distributions of the size of each digit look approximately similar, the modes of certain digits  in $\score-$MNIST are closer (like $1,5,6,7$) whereas the other digits are more separated ($2,3,9$) -- this slight variation is captured nicely by our $F_1$ metric regularized models, thereby preserving the accuracy to a larger extent than the baseline.

\section{Why Choose the Category ``Sky'' in MSCOCO}
\begin{figure*}[!h]
	\centering\label{fig:sky}
	\begin{subfigure}[!t]{0.45\linewidth}
		{\includegraphics[width=1\linewidth]{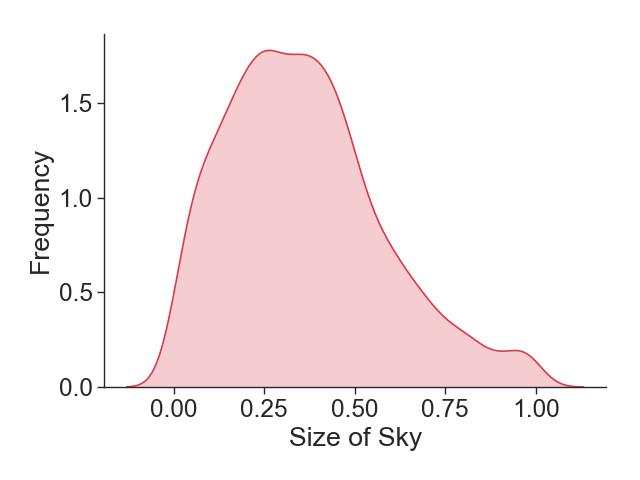}}
	\end{subfigure}
	\begin{subfigure}[!t]{0.45\linewidth}
		{\includegraphics[width=1\linewidth]{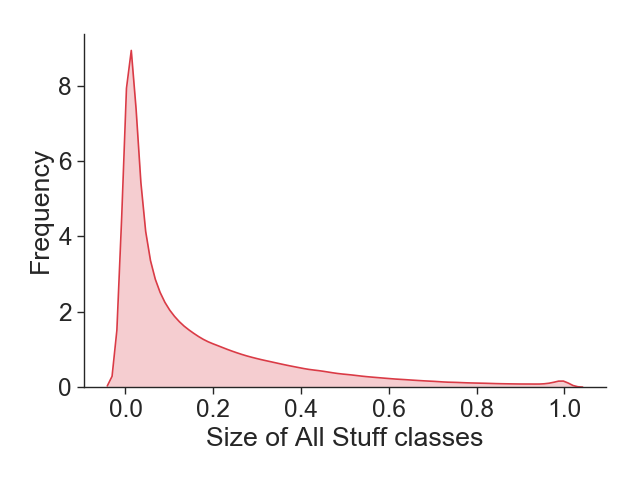}}
	\end{subfigure}
\caption{Histograms of Sky vs. Typical class in MS-COCO dataset}\label{fig:cocosky}
\end{figure*}

Essentially, we observed that images corresponding to the class sky has the most amount of variance with respect to size as shown in Figure \ref{fig:cocosky}. In Figure \ref{fig:cocosky}, the histogram of the Sky class is shown on the left with a typical class from MS-COCO dataset shown on the right. Notice that the tail of the Sky class is significantly fatter than that of a typical class. Even though the Sky class is bimodal, our experimental results in the main paper show that nondecomposable regularizers can provide significant benefits for classes with fat tails.

\section{Additional Results on MS-COCO Full Dataset}  We provide the quantitative comparison of baseline and our algorithm mentioned in the main paper.
\begin{figure}[!t]
	\begin{subfigure}[!t]{1\linewidth}
		\includegraphics[width=1\linewidth]{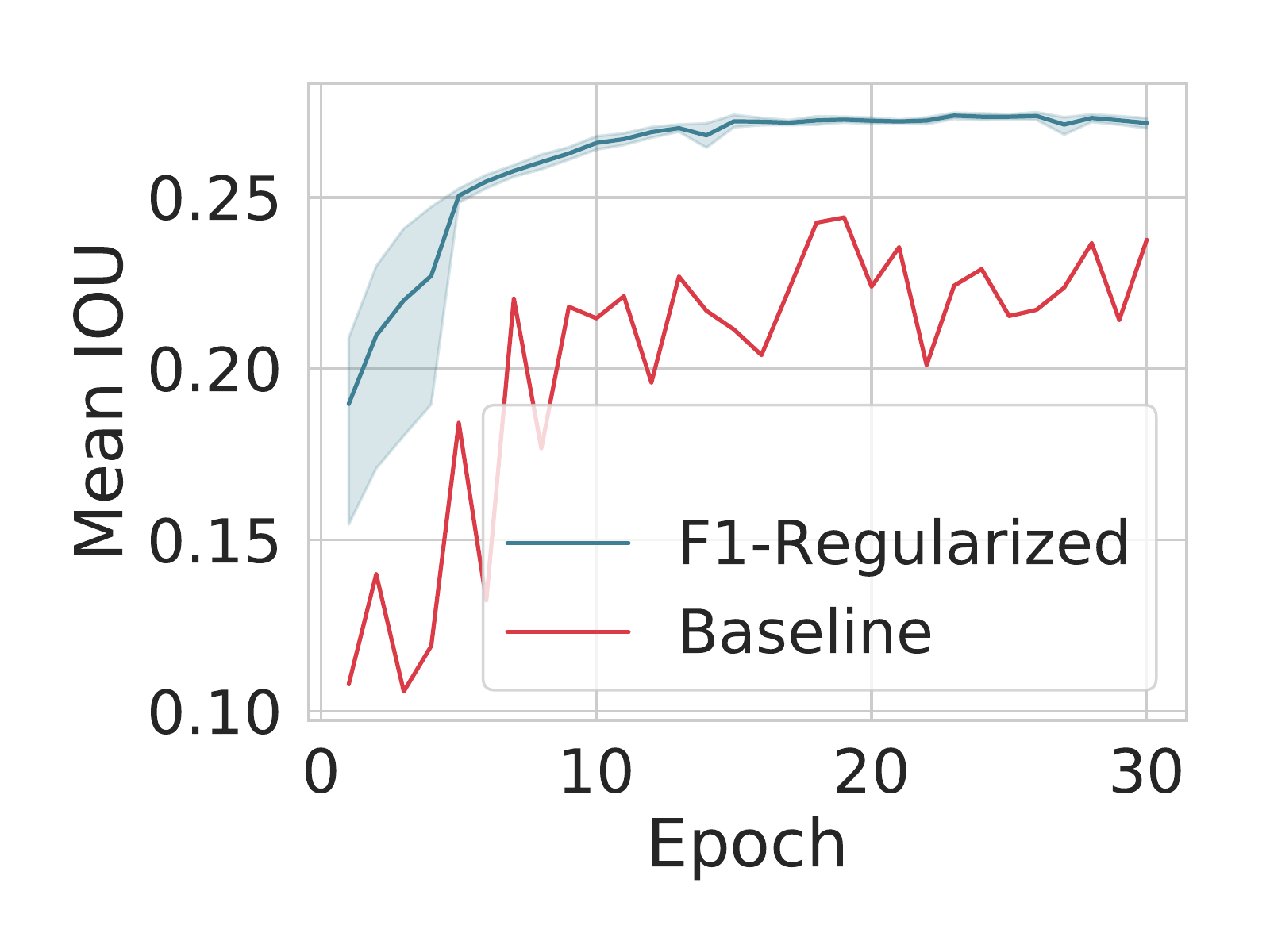}
	\end{subfigure}			
	\caption{\label{fig:mscoco164k} Quantitative results on MSCOCO stuff segmentation benchmark. Our mean IOU improves upon the state of the art
		reported in \cite{chen2018encoder} by $\approx 10\%$. We outperform the baseline on MSCOCO 164K while being stable;} 
\end{figure}
 Now we provide some more qualitative results of our Algorithm 1 on the full MS-COCO dataset which consists of $164k$ images to substantiate our claim that both small/large sized objects are segmented equally well when models are regularized with data dependent nondecomposable regularizers.

\begin{figure*}[!t]
	\centering
	
	\begin{subfigure}[!t]{0.24\linewidth}
		{\includegraphics[width=1\linewidth]{images/full_coco/images/000000011760.jpg}}
	\end{subfigure}
	\begin{subfigure}[!t]{0.24\linewidth}
		{\includegraphics[width=1\linewidth]{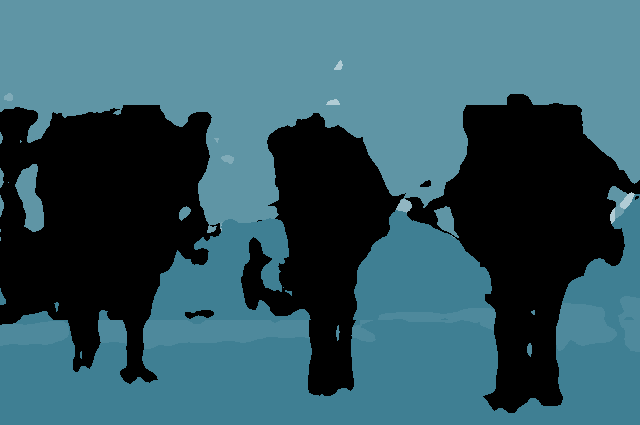}}
	\end{subfigure}
	\begin{subfigure}[!t]{0.24\linewidth}
		{\includegraphics[width=1\linewidth]{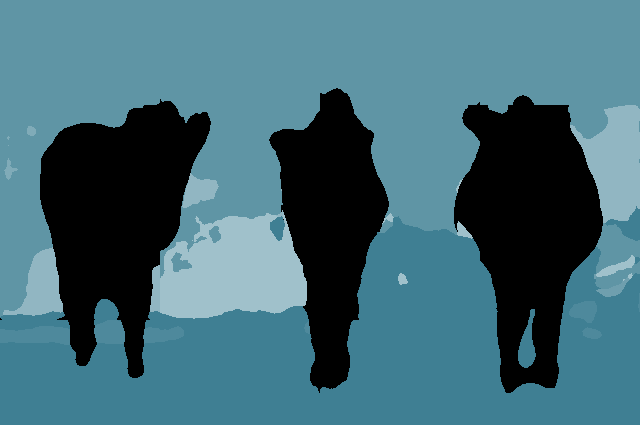}}
	\end{subfigure}
	\begin{subfigure}[!t]{0.24\linewidth}
		{\includegraphics[width=1\linewidth]{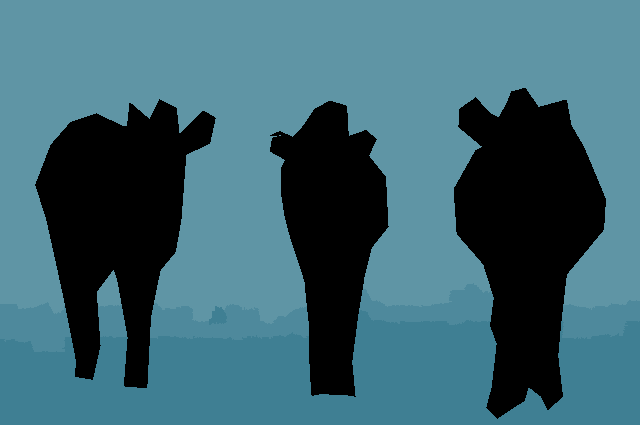}}
	\end{subfigure}	
	\begin{subfigure}[!t]{0.24\linewidth}
		{\includegraphics[width=1\linewidth]{images/full_coco/images/000000025057.jpg}}
	\end{subfigure}
	\begin{subfigure}[!t]{0.24\linewidth}
		{\includegraphics[width=1\linewidth]{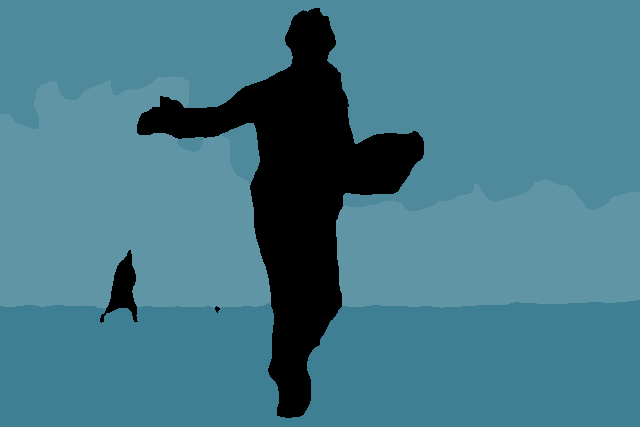}}
	\end{subfigure}
	\begin{subfigure}[!t]{0.24\linewidth}
		{\includegraphics[width=1\linewidth]{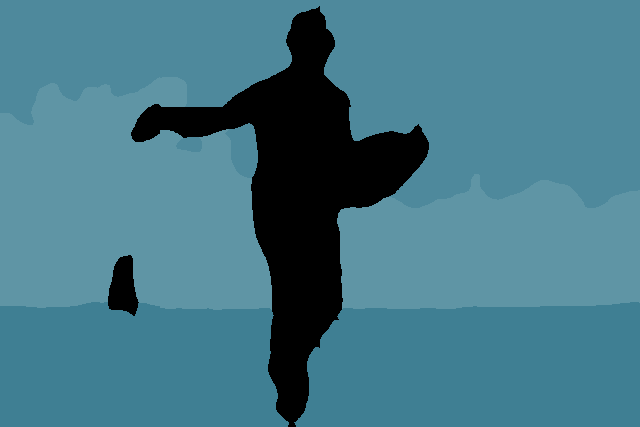}}
	\end{subfigure}
	\begin{subfigure}[!t]{0.24\linewidth}
		{\includegraphics[width=1\linewidth]{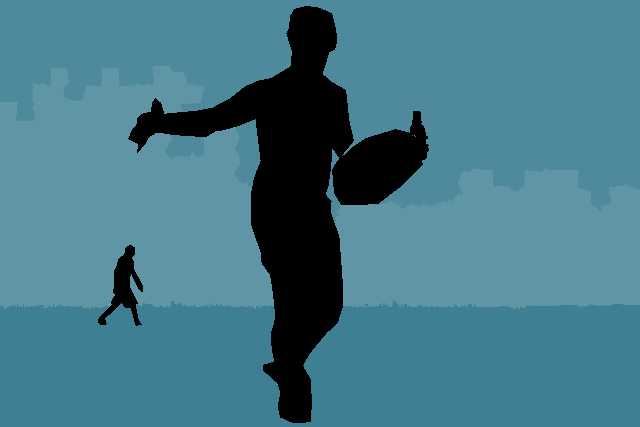}}
	\end{subfigure}	

	\begin{subfigure}[!t]{0.24\linewidth}
		{\includegraphics[width=1\linewidth]{images/full_coco/images/000000170545.jpg}}
	\end{subfigure}
	\begin{subfigure}[!t]{0.24\linewidth}
		{\includegraphics[width=1\linewidth]{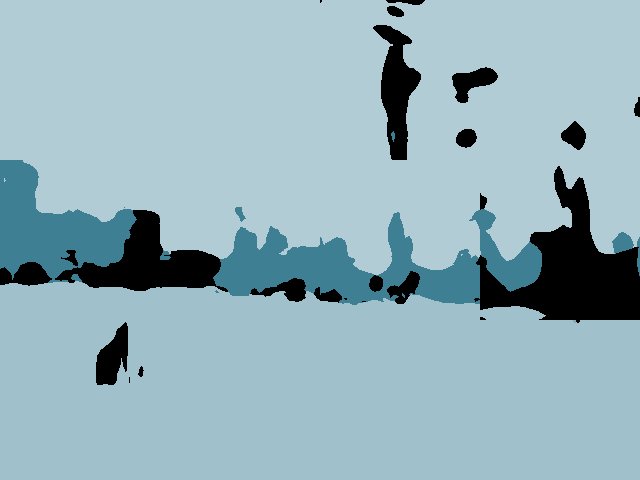}}
	\end{subfigure}
	\begin{subfigure}[!t]{0.24\linewidth}
		{\includegraphics[width=1\linewidth]{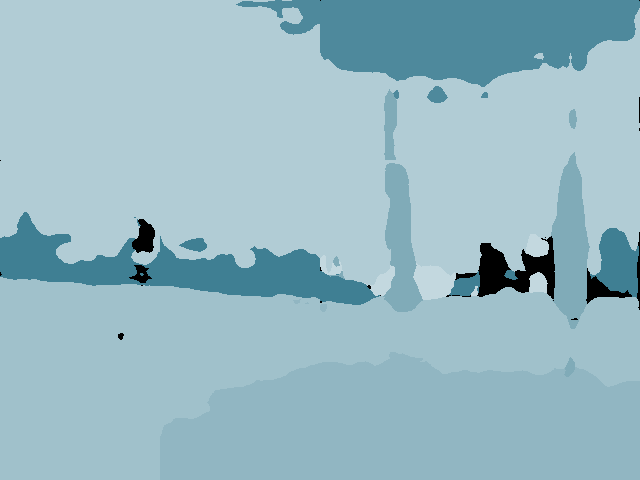}}
	\end{subfigure}
	\begin{subfigure}[!t]{0.24\linewidth}
		{\includegraphics[width=1\linewidth]{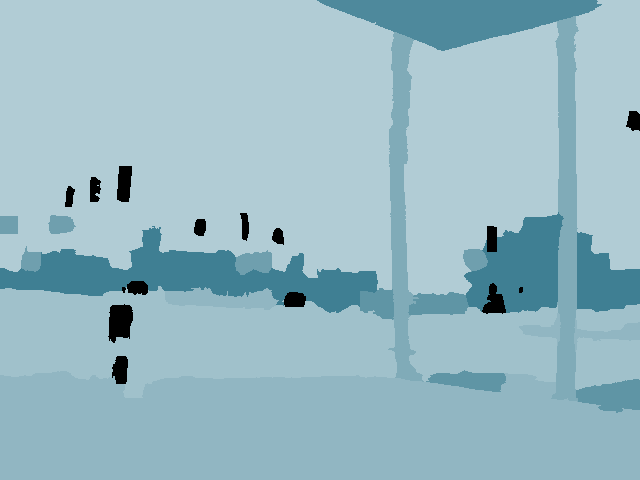}}
	\end{subfigure}

	\begin{subfigure}[!t]{0.24\linewidth}
		{\includegraphics[width=1\linewidth]{images/full_coco/images/000000511599.jpg}}
	\end{subfigure}
	\begin{subfigure}[!t]{0.24\linewidth}
		{\includegraphics[width=1\linewidth]{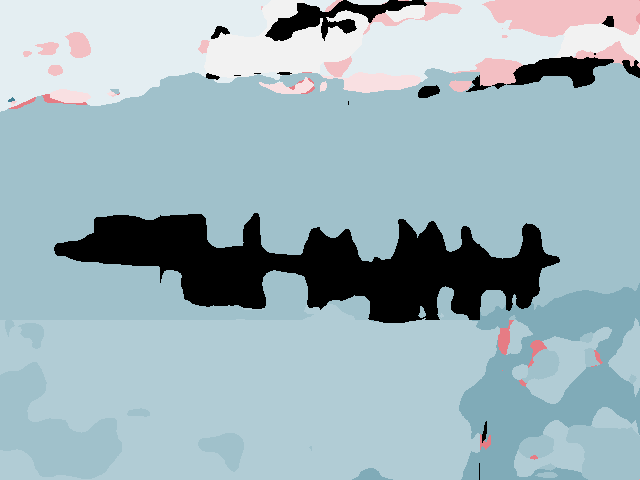}}
	\end{subfigure}
	\begin{subfigure}[!t]{0.24\linewidth}
		{\includegraphics[width=1\linewidth]{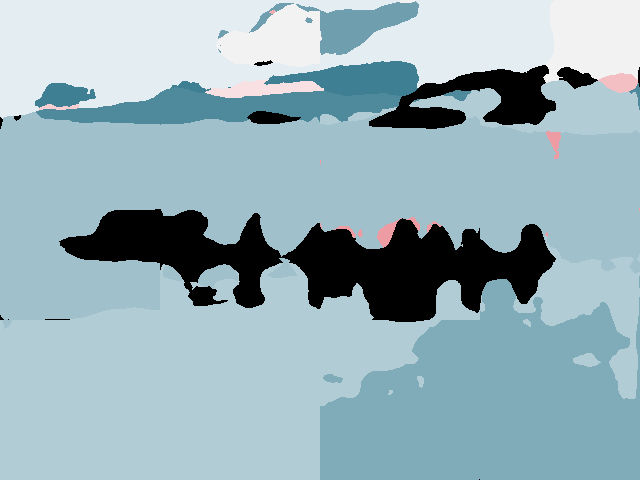}}
	\end{subfigure}
	\begin{subfigure}[!t]{0.24\linewidth}
		{\includegraphics[width=1\linewidth]{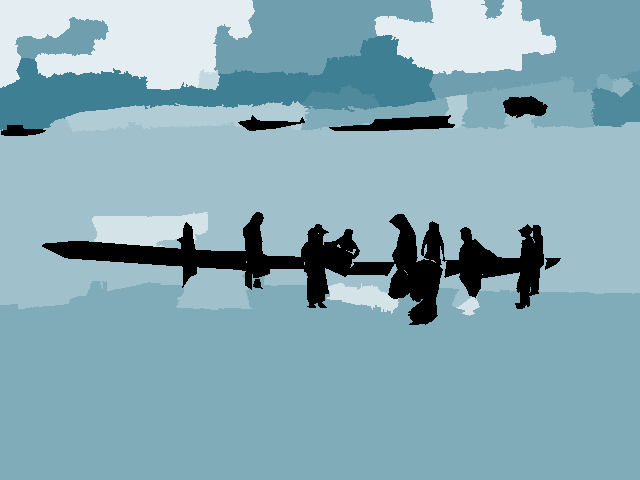}}
	\end{subfigure}	\caption{From left to right: image, baseline (DeepLabV3+ without F1-normalization), our method, and the ground truth. Our method qualitatively performs better than the baseline. Our regularizer is able to "smoothen" the masks, thereby improving the quality and removing many artifacts from the images.}	
		\end{figure*}
\begin{figure*}
	\begin{subfigure}[!t]{0.24\linewidth}
		{\includegraphics[width=1\linewidth]{images/full_coco/images/000000276024.jpg}}
	\end{subfigure}
	\begin{subfigure}[!t]{0.24\linewidth}
		{\includegraphics[width=1\linewidth]{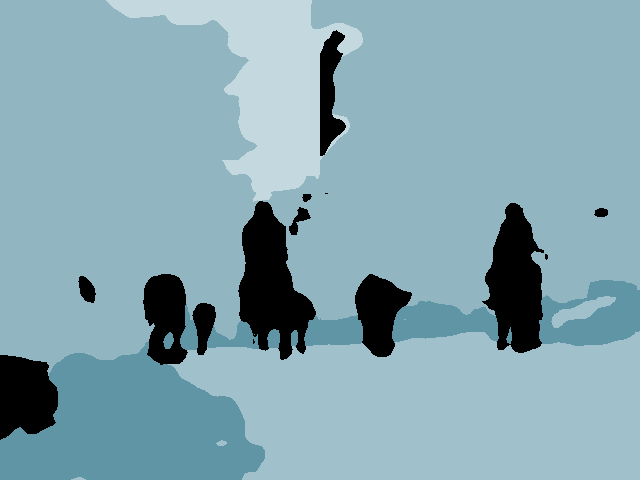}}
	\end{subfigure}
	\begin{subfigure}[!t]{0.24\linewidth}
		{\includegraphics[width=1\linewidth]{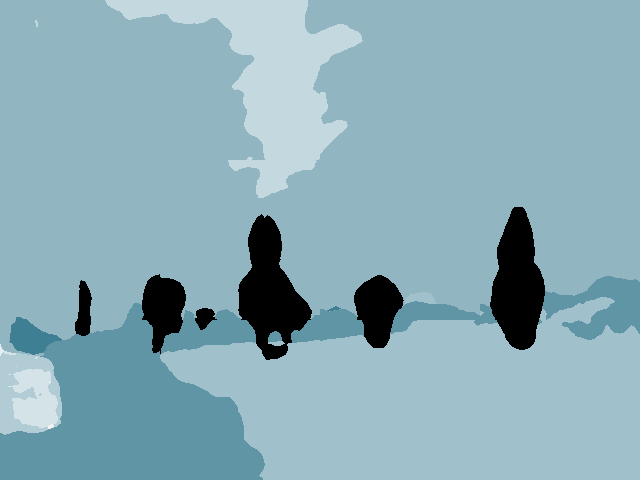}}
	\end{subfigure}
	\begin{subfigure}[!t]{0.24\linewidth}
		{\includegraphics[width=1\linewidth]{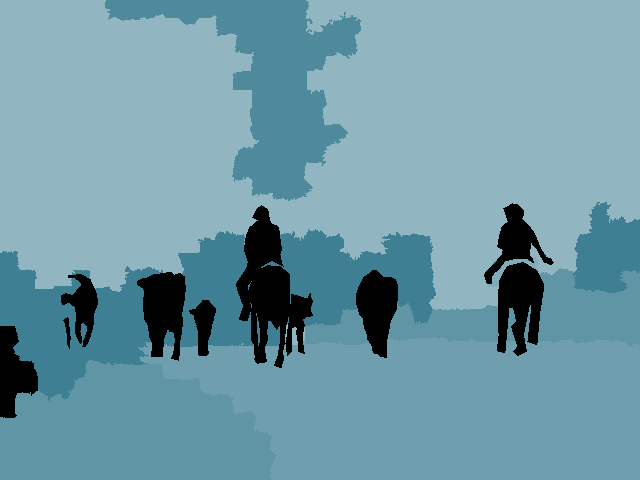}}
	\end{subfigure}		

	\begin{subfigure}[!t]{0.24\linewidth}
		{\includegraphics[width=1\linewidth]{images/full_coco/images/000000275727.jpg}}
	\end{subfigure}
	\begin{subfigure}[!t]{0.24\linewidth}
		{\includegraphics[width=1\linewidth]{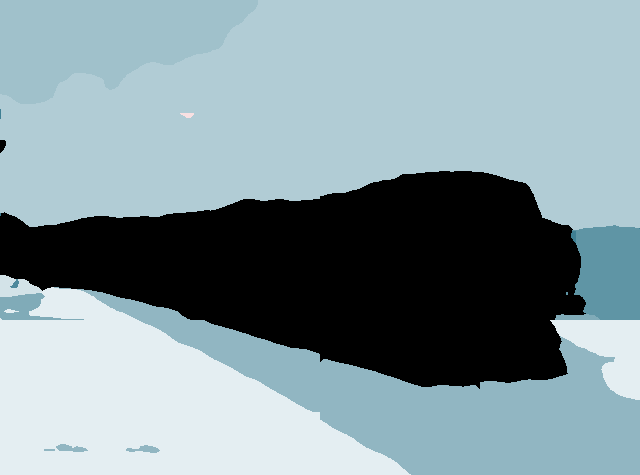}}
	\end{subfigure}
	\begin{subfigure}[!t]{0.24\linewidth}
		{\includegraphics[width=1\linewidth]{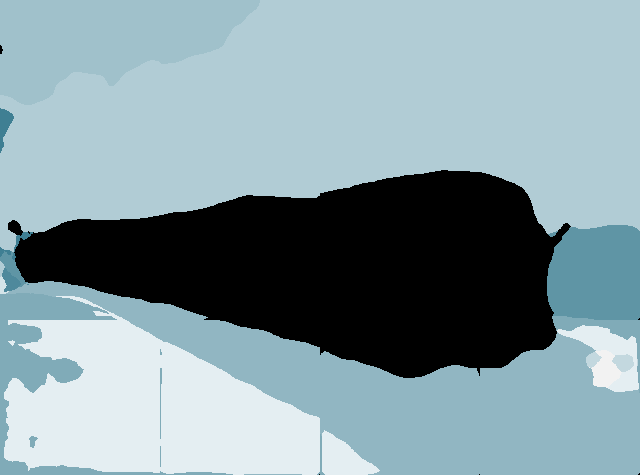}}
	\end{subfigure}
	\begin{subfigure}[!t]{0.24\linewidth}
		{\includegraphics[width=1\linewidth]{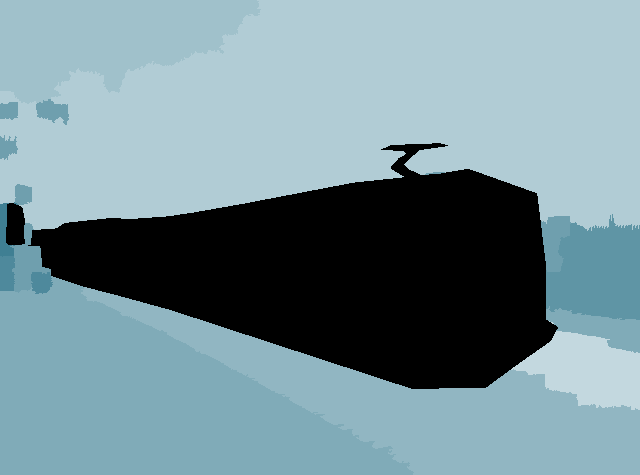}}
	\end{subfigure}	

	\begin{subfigure}[!t]{0.24\linewidth}
		{\includegraphics[width=1\linewidth]{images/full_coco/images/000000147725.jpg}}
	\end{subfigure}
	\begin{subfigure}[!t]{0.24\linewidth}
		{\includegraphics[width=1\linewidth]{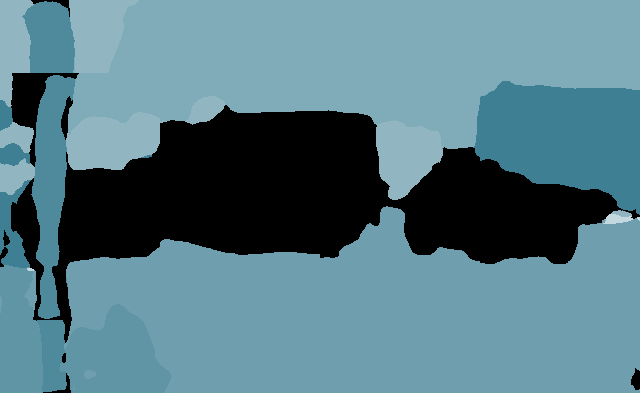}}
	\end{subfigure}
	\begin{subfigure}[!t]{0.24\linewidth}
		{\includegraphics[width=1\linewidth]{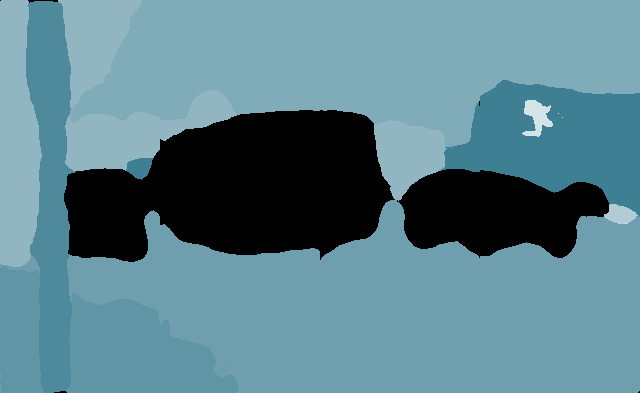}}
	\end{subfigure}
	\begin{subfigure}[!t]{0.24\linewidth}
		{\includegraphics[width=1\linewidth]{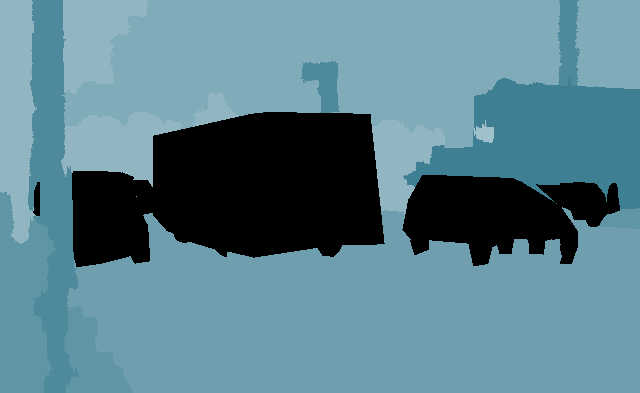}}
	\end{subfigure}	

	\caption{From left to right: image, baseline (DeepLabV3+ without F1-normalization), our method, and the ground truth. Our method qualitatively performs better than the baseline. Our regularizer is able to "smoothen" the masks, thereby improving the quality and removing many artifacts from the images.}
\end{figure*}

\begin{figure*}[!t]
	\centering
	\begin{subfigure}[!t]{0.3\linewidth}
		{\includegraphics[width=1\linewidth]{images/coco_sky_examples_and_hist/000000024027.jpg}}
	\end{subfigure}
	\begin{subfigure}[!t]{0.3\linewidth}
		{\includegraphics[width=1\linewidth]{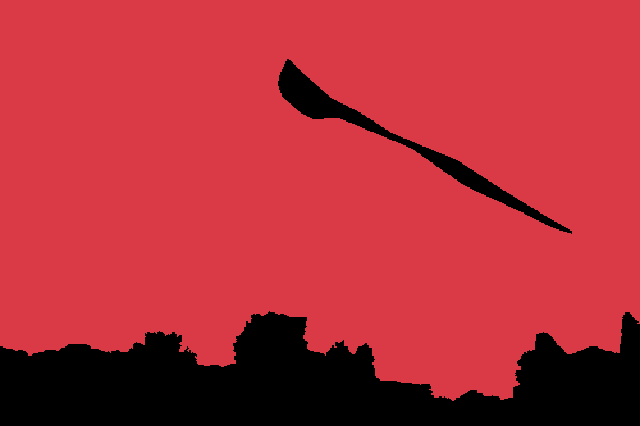}}
	\end{subfigure}
	\begin{subfigure}[!t]{0.3\linewidth}
		{\includegraphics[width=1\linewidth]{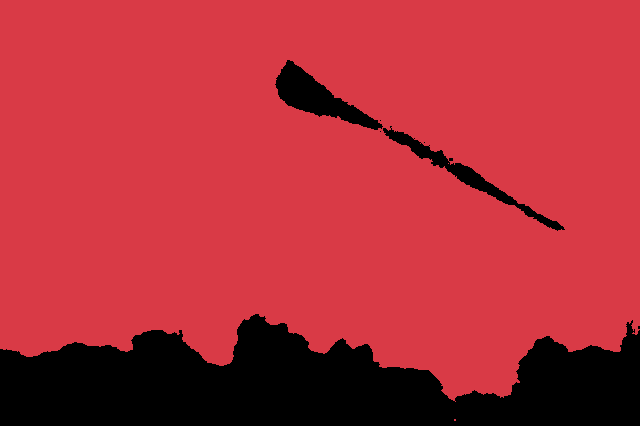}}
	\end{subfigure}
	\begin{subfigure}[!t]{0.3\linewidth}
		{\includegraphics[width=1\linewidth]{images/coco_sky_examples_and_hist/000000021903.jpg}}
	\end{subfigure}
	\begin{subfigure}[!t]{0.3\linewidth}
		{\includegraphics[width=1\linewidth]{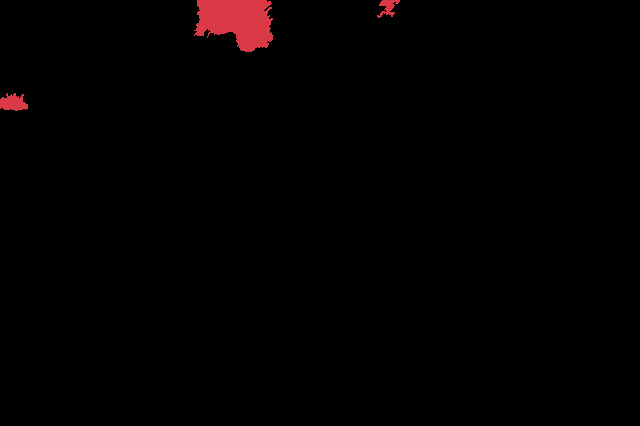}}
	\end{subfigure}
	\begin{subfigure}[!t]{0.3\linewidth}
		{\includegraphics[width=1\linewidth]{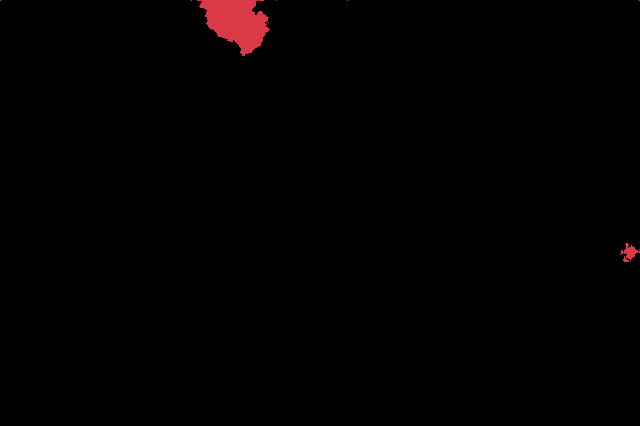}}
	\end{subfigure}
	\begin{subfigure}[!t]{0.3\linewidth}
		{\includegraphics[width=1\linewidth]{images/coco_sky_examples_and_hist/000000161879.jpg}}
	\end{subfigure}
	\begin{subfigure}[!t]{0.3\linewidth}
		{\includegraphics[width=1\linewidth]{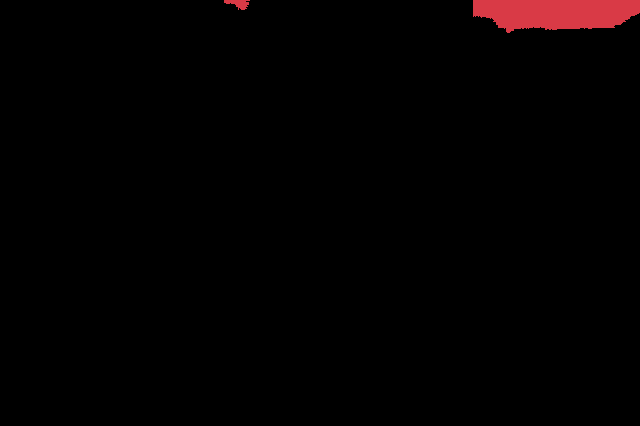}}
	\end{subfigure}
	\begin{subfigure}[!t]{0.3\linewidth}
		{\includegraphics[width=1\linewidth]{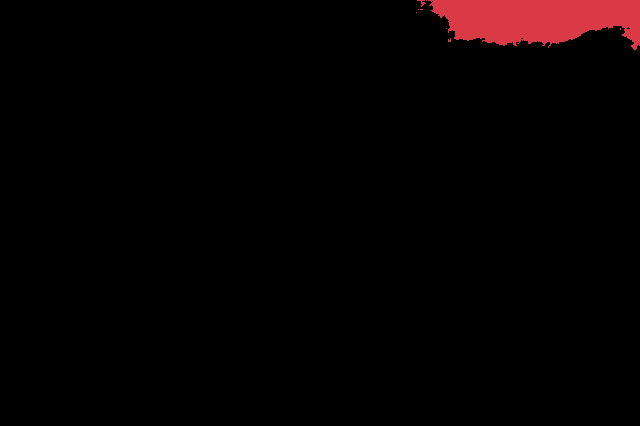}}
	\end{subfigure}
	\begin{subfigure}[!t]{0.3\linewidth}
		{\includegraphics[width=1\linewidth]{images/coco_sky_examples_and_hist/000000323571.jpg}}
	\end{subfigure}
	\begin{subfigure}[!t]{0.3\linewidth}
		{\includegraphics[width=1\linewidth]{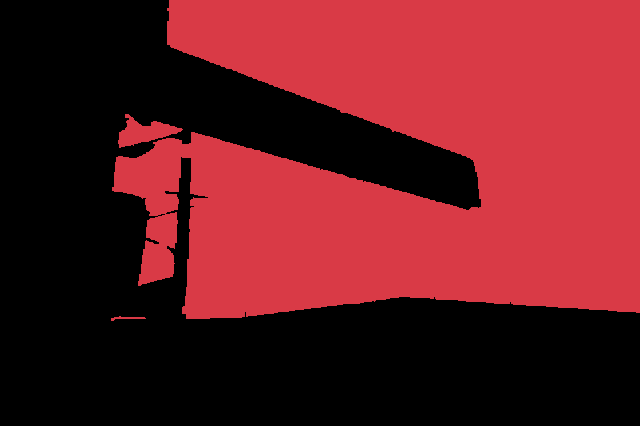}}
	\end{subfigure}
	\begin{subfigure}[!t]{0.3\linewidth}
		{\includegraphics[width=1\linewidth]{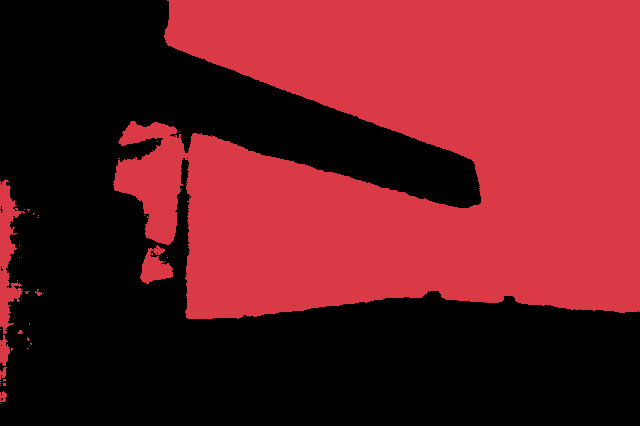}}
	\end{subfigure}
	\caption{From left to right: image, baseline (DeepLabV3+ without F1-normalization), our method, and the ground truth. Our method qualitatively performs better than the baseline. Our regularizer is able to "smoothen" the masks, thereby improving the quality and removing many artifacts from the images.}
\end{figure*}
\begin{figure*}
	\begin{subfigure}[!t]{0.3\linewidth}
		{\includegraphics[width=1\linewidth]{images/coco_sky_examples_and_hist/000000002532.jpg}}
	\end{subfigure}
	\begin{subfigure}[!t]{0.3\linewidth}
		{\includegraphics[width=1\linewidth]{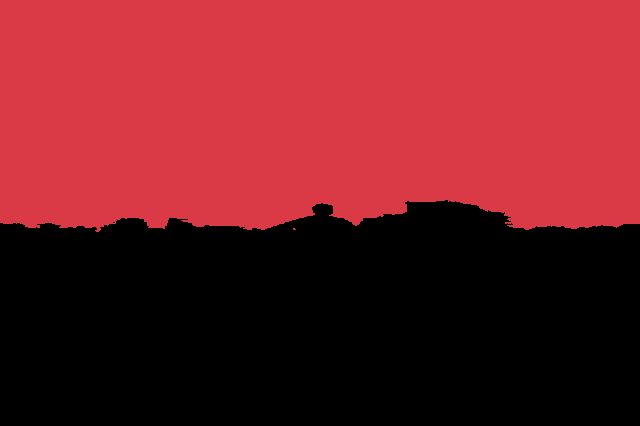}}
	\end{subfigure}
	\begin{subfigure}[!t]{0.3\linewidth}
		{\includegraphics[width=1\linewidth]{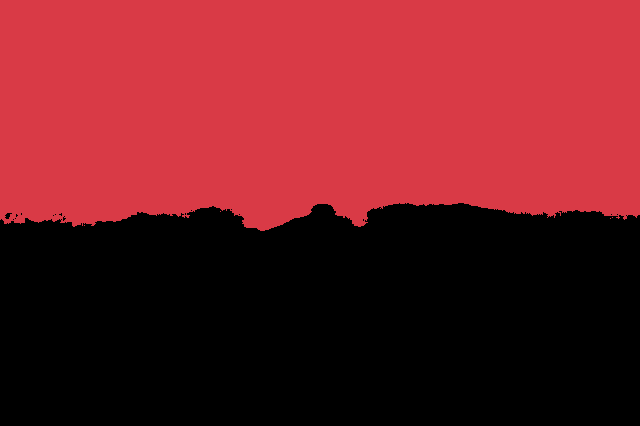}}
	\end{subfigure}
	
	\begin{subfigure}[!t]{0.3\linewidth}
		{\includegraphics[width=1\linewidth]{images/coco_sky_examples_and_hist/000000013291.jpg}}
	\end{subfigure}
	\begin{subfigure}[!t]{0.3\linewidth}
		{\includegraphics[width=1\linewidth]{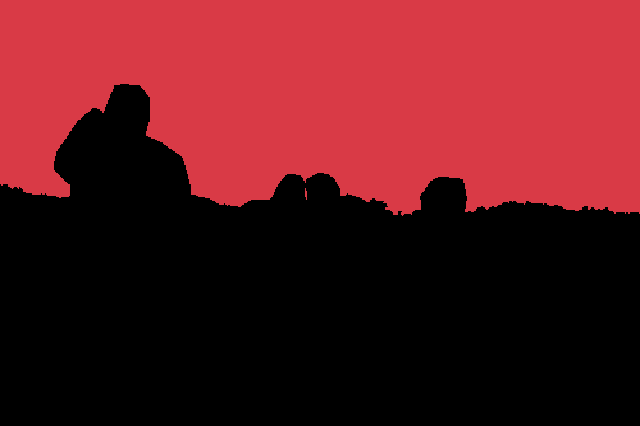}}
	\end{subfigure}
	\begin{subfigure}[!t]{0.3\linewidth}
		{\includegraphics[width=1\linewidth]{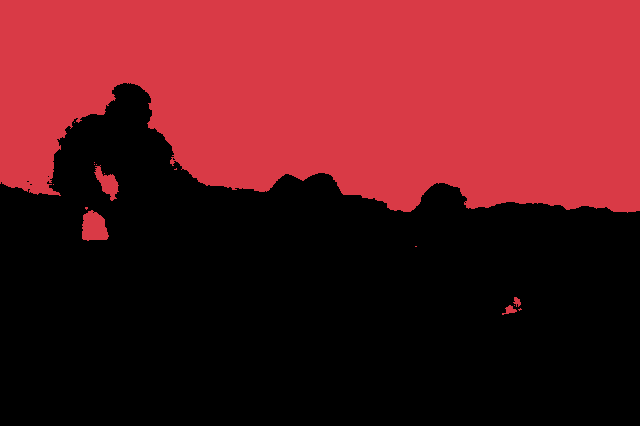}}
	\end{subfigure}
	
	\begin{subfigure}[!t]{0.3\linewidth}
		{\includegraphics[width=1\linewidth]{images/coco_sky_examples_and_hist/000000166391.jpg}}
	\end{subfigure}
	\begin{subfigure}[!t]{0.3\linewidth}
		{\includegraphics[width=1\linewidth]{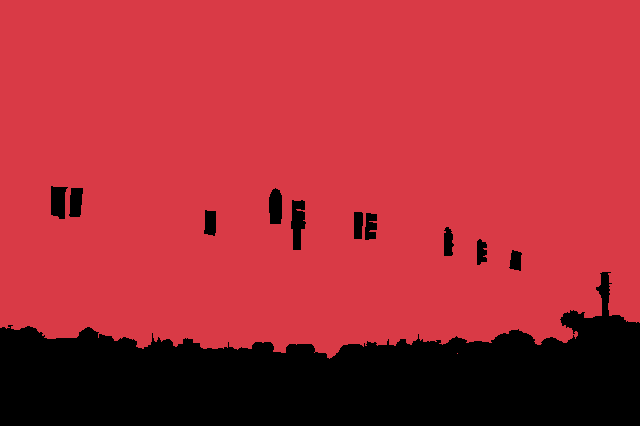}}
	\end{subfigure}
	\begin{subfigure}[!t]{0.3\linewidth}
		{\includegraphics[width=1\linewidth]{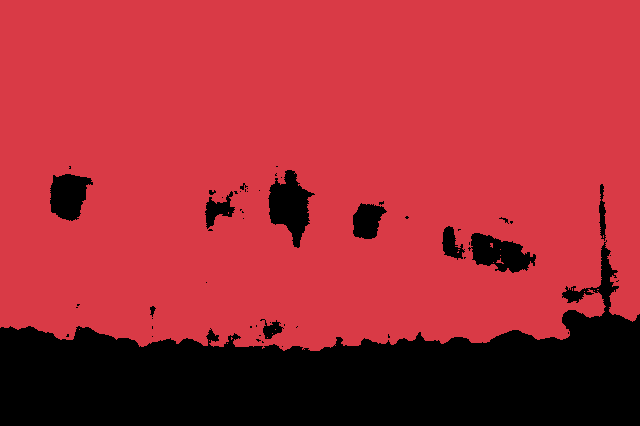}}
	\end{subfigure}
	\begin{subfigure}[!t]{0.3\linewidth}
		{\includegraphics[width=1\linewidth]{images/coco_sky_examples_and_hist/000000032941.jpg}}
	\end{subfigure}
	\begin{subfigure}[!t]{0.3\linewidth}
		{\includegraphics[width=1\linewidth]{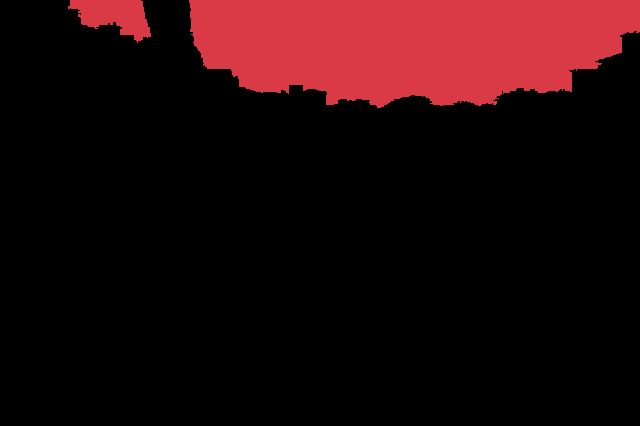}}
	\end{subfigure}
	\begin{subfigure}[!t]{0.3\linewidth}
		{\includegraphics[width=1\linewidth]{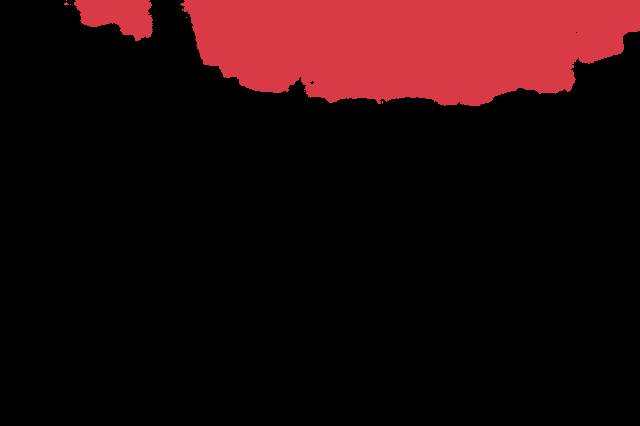}}
	\end{subfigure}
	\caption{From left to right: image, baseline (DeepLabV3+ without F1-normalization), our method, and the ground truth. Our method qualitatively performs better than the baseline. Our regularizer is able to "smoothen" the masks, thereby improving the quality and removing many artifacts from the images.}
\end{figure*}

\end{document}